June 5, 2019

# Visual Backpropagation

**Roy S. Freedman**[1]

**Abstract**

We show how a declarative functional programming specification of backpropagation yields a visual and transparent implementation within spreadsheets. We call our method "Visual Backpropagation." This backpropagation implementation exploits array worksheet formulas, manual calculation, and has a sequential order of computation similar to the processing of a systolic array. The implementation uses no hidden macros nor user-defined functions; there are no loops, assignment statements, or links to any procedural programs written in conventional languages. As an illustration, we compare a Visual Backpropagation solution to a Tensorflow (Python) solution on a standard regression problem.

## 1 Introduction

Neural networks are nonlinear formulas that approximate functions. Backpropagation [1] is a machine learning algorithm that creates these approximation formulas iteratively. The iterative approach has been likened to finding the top of a mountain in a mountain range: at every iteration, we take a step in the direction of the steepest slope. We stop taking steps when the approximation is "good enough." Mathematicians call this approach "gradient search."

Backpropagation is a proven and popular machine learning technique since its development over thirty years ago. According to LeCun et al [2], backpropagation is popular because it is "conceptionally simple, computationally efficient, and because it often works." Bishop [3, p.246] concurs: "One of the most important aspects of backpropagation is its computational efficiency."

Backpropagation has been programmed in C, C++, Fortran, Java, Julia, Python, R, Visual Basic, and other procedural (imperative) languages. Procedural programs implementations rely on constructs like loops and assignment statements. Programs typically result in text that looks quite different and more complicated than the actual mathematical specification. (My favorite C implementation of backpropagation is the 17 pages of C code in Pao [4, Appendix A]). In general, these programs are not for neophytes.

Getting a backpropagation-based machine learning experiment to actually work (meaning – in a technical sense – to converge to a meaningful solution) has been described as more than an art than a science [2]. The reason for this is the number of design and parameter choices that need to be made. These choices include specifying the nature of the approximating function (topology, activation functions); specifying properties of the iteration (learning rates); specifying techniques that assist convergence (data scaling); specifying how the answer will be evaluated (data selection for in-sample training and out-sample tuning and cross-validation); and, finally, specifying the criteria used to determine when to stop (when the approximation is "good enough" based on determining average errors).

[1] Roy S. Freedman is with Inductive Solutions, Inc., New York, NY 10280 and with the Department of Finance and Risk Engineering, New York University Tandon School of Engineering, Brooklyn NY 11201. Email: roy@inductive.com.

Current backpropagation implementations – especially those used for academic and professional education – rely on open source software languages and libraries. All design and parameter choices need to be programmed. Some implementations graft a front-end user interface – such as a spreadsheet – to a hidden backpropagation back-end engine. The interface manages data collection and parameter selection via an Application Programming Interface, Dynamic Link Library, or other inter-process communication scheme. Etheridge and Brooks list a number of such vendors in their 1994 paper [5].

The motivation of this paper is to explain the purpose and use of backpropagation to students and professionals in finance by using spreadsheets as a transparent backpropagation engine. The target audience includes non-experts in machine learning and artificial intelligence; yet, as finance professionals, they are familiar with linear regression and spreadsheets. We show how the backpropagation algorithm can be implemented in Excel spreadsheets using Excel worksheet functions like array and matrix multiplication. We call our method "Visual Backpropagation." We use "pure Excel" – there are no dynamic link libraries to C, C++, C#, Java, Python, Visual Basic for Applications (VBA), or any other language. There are no macros no user-defined functions and no subroutines; there are no loops, assignment statements, or any procedural programming constructs.

In Section 2 we first review the array-based and spreadsheet-influenced notation used in Visual Backpropagation. We also show how neural network nonlinear formulas generalize the formulas used in linear regression. Section 3 reviews the semantics of spreadsheet computation – especially computation under the "manual mode of calculation" and specifies our visual backpropagation formulas. Here we use a simple problem (the nonlinearly separable "exclusive or" together with "and") as an example for a 2-hidden layer learning system (2-2-2-1 topology). (Additional details associated with manual calculation are in the Appendix.) Section 4 shows Visual Backpropagation solving a practical problem (the Auto Mpg problem found in the UCI Machine Learning repository [6]). This example is used in the Tensorflow regression tutorial [7]). Finally, Section 5 summarizes our work and offers some proposals for future work and extensions.

Since our approach is visual, we provide many figures showing screen images that illustrate the Visual Backpropagation method. We preface figure captions by section number. The actual working spreadsheets (content covered by the Creative Commons Attribution 3.0 License, and formulas by the Apache 2.0 License) are posted on a website [8].

## Why spreadsheets?

Spreadsheets are the lingua franca of the computing world of accounting, economics, finance, and basic brute-force data collecting for data sets of less than a million records. For most people, spreadsheets are easy to learn and easy to use. There are dozens of spreadsheet implementations (some are open source) but even after several decades, Microsoft Excel is the most popular spreadsheet platform. Walsh [9] summarizes seven reasons why at least half a billion people use Excel. Despite this, the spreadsheet programming paradigm has been ignored by academic computer science and many programming professionals: "Real programmers don't use spreadsheets" [10]. One reason is that

> The neglect of spreadsheets in the programming language literature might be caused by an aversion to a language created by students who had the bad taste of using their ideas to earn money instead of starting a Computing Science career that might have led to a Turing award. I don't think, however, that computer language specialists harbor such spiteful thoughts, and therefore, I suggest that spreadsheets are intrinsically uninteresting. Presumably, many others who have looked at spreadsheets have come to

> the same conclusion, and such pedestrian results have never been seen fit for publication (or at least, they have not been accepted by journal or conference referees)[10].

Nevertheless, the same author admits

> In spreadsheets the ideas from two powerful languages, LISP and APL are combined. The acceptance of spreadsheets by non-programmers shows that functional programming and the use of arrays as the only data structure are concepts that are easy to understand [10].

The academic disdain for spreadsheets continues. Indeed, Fouhey and Maturana [11] posted a "joke paper" that mocks a fake use of Excel in connection with machine learning. But Walsh [9] concludes:

> …mention Excel to techies and it's often dismissed with a sniff. However, somewhat "like the love that dare not speak its name", the vast majority of users in the business world use Excel and practically every system has a button that says "Give it to me in Excel".

This situation is changing. In 2014, in a serious and rigorous academic monograph of spreadsheets, Sestoff [12] showed that spreadsheets really are dynamically typed functional programs. Functional programs are declarative: these programs do not rely on procedural constructs. Declarative functional programs look like mathematical specifications. In a sense, spreadsheet programming consists of (1) selecting and naming ranges of cells, and (2) entering (or re-entering) values and formulas in cells. Program testing consists of tracing the propagation of values via formula computation from cell to cell or region to region. Spreadsheet programming declaratively specifies names and computational expressions. In contrast, programming in conventional imperative languages specifies a schedule of control flow with procedural statements.

## 2 Array Notations for Backpropagation

This section specifies our notation for backpropagation in the context of linear algebra (i.e., matrix and tensor operations) and show how these operations have an equivalence to Excel array operations. We do not derive the backpropagation algorithm; there are plenty of nice derivations, in particular, see Bishop [3] or Pao [4]; Efron and Hastie [13, p. 356-9] use a vector notation almost similar to ours. We re-specify the backpropagation algorithm in a form suitable for spreadsheet formula computation.

We assume the reader is familiar with basic spreadsheet concepts:

- How spreadsheets reference cells, and how spreadsheets define named cell regions;
- How formulas, array formulas, and built-in functions are inserted into cells;
- How formula-derived values propagate computation across cells;
- How computation is achieved via calculation options (automatic vs. manual).

### 2.1 Review of Array Operations

Array operations originate from the mathematics of vectors, matrices, and tensors. We show how the many operations on these mathematical objects are modeled by Excel array operations. We use our formalism to map backpropagation into a tensor form and ultimately to Excel array equivalences – as copy-and-paste array formulas.

We first start with a notation for naming. Spreadsheet arrays are rectangular regions of cells (or a single cell). Denote ranges by bold type. Here is a formal definition of two arrays: one has three values in its rows and the other has two values in its columns (sometimes called a row array and column array, respectively):

$$\mathbf{a}:(1\times 3) \text{ and } \mathbf{b}:(2\times 1) \ .$$

In Excel spreadsheet notation, instead of bold type for naming arrays, we use plain type (frequently with an underscore to make sure our name is not confused with a row or a column location; for example, the name "a2" denotes the cell in the second row and first column of a worksheet). Many spreadsheets have a Define Name command that binds names to spreadsheet arrays. For example, the Define Name in Excel defines array a_ via

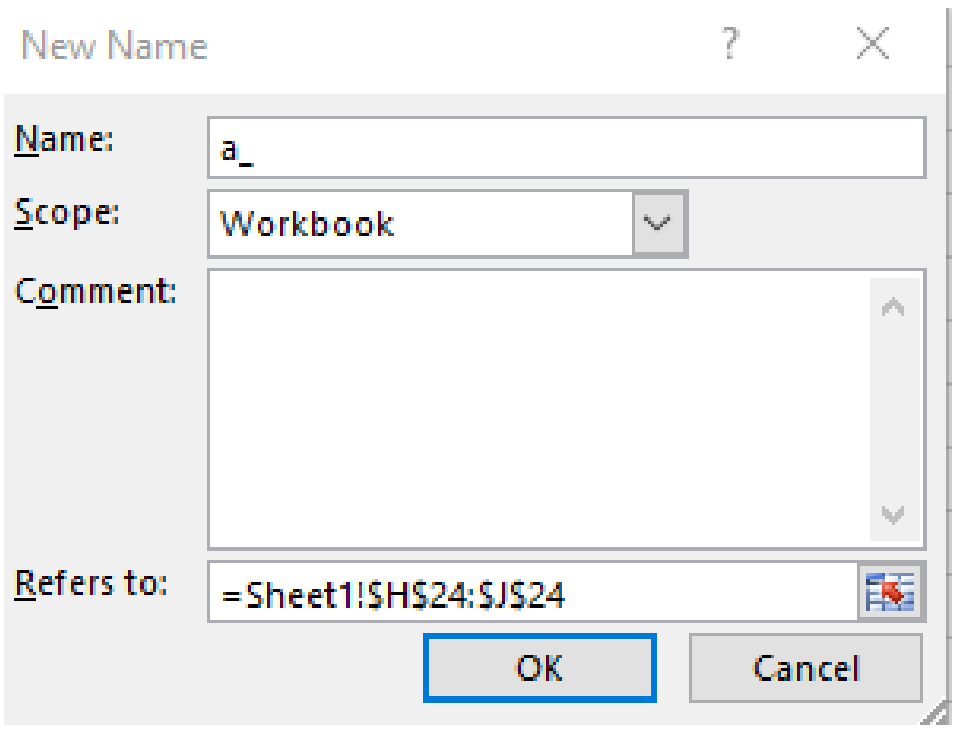

This defines array a_ as a place on the worksheet.

Extend these definitions to names of components by specifying values. We use ":=" to denote "evaluates to" (for names and formulas) and "=" to denote "the same as" (when checking equality of expressions). The individual cells in an array are called components (indicated in non-bold type). The standard convention is to use subscripts to denote row-column locations. For example, is the component in the first row and second column; $a_{12}$ is the component in the second row and first column:

$$\mathbf{a}:(1\times 3):=\begin{pmatrix} a_{11} & a_{12} & a_{13} \end{pmatrix} \text{ and } \mathbf{b}:(2\times 1):=\begin{pmatrix} b_{11} \\ b_{21} \end{pmatrix} \ .$$

Excel does this with the OFFSET worksheet function:

$$\text{=OFFSET(a_, 0,1,1,1)} \ := \ a_{12}$$

To avoid ambiguity, the definition includes the number of rows and the number of columns via $(row \times column)$. Frequently we call the number of rows (of a column array) or number of columns (of a row array) the dimension.

The transpose of an array is an operation that switches the rows and columns:

$$\mathbf{a}^T:(1\times 3):=\left(\mathbf{a}:(3\times 1)\right)^T:=\begin{pmatrix} a_{11} & a_{12} & a_{13} \end{pmatrix}^T:=\begin{pmatrix} a_{11} \\ a_{12} \\ a_{13} \end{pmatrix}$$

The transpose operations changes an n-dimensional row vector to an n-dimensional column vector (and vice versa!). In notational shorthand: $((n\times 1):)^T = (1\times n):$ . Excel does this with the TRANSPOSE worksheet function.

We can specify a new column array by appending a row to an existing column array. For example, an extension of vector $\mathbf{b}$ is new column vector $\mathbf{b}^*$ that appends a row below the last row of $\mathbf{b}$. Suppose this row has value 1. Specify this extension by:

$$\mathbf{b}^*:(3\times 1) := \begin{pmatrix} \mathbf{b}:(2\times 1) \\ 1 \end{pmatrix} := \begin{pmatrix} \begin{pmatrix} b_{11} \\ b_{21} \end{pmatrix} \\ 1 \end{pmatrix} := \begin{pmatrix} b_{11} \\ b_{21} \\ 1 \end{pmatrix}$$

Note that there is nothing special about 1: the added value can be any number. Similarly, denote a restriction of column array $\mathbf{a}$ by removing the last column; the restriction of a row array removes the last row. For example:

$$\begin{aligned} &(\mathbf{a}:(1\times 3)):(1\times 2) := \mathbf{a}:(1\times 2) := \begin{pmatrix} a_{11} & a_{12} \end{pmatrix} \\ &\mathbf{b}^*:(2\times 1) := \mathbf{b} \end{aligned} .$$

In this sense, the $(row \times column)$ specification becomes an operator for a restriction. (In linear algebra, single column arrays are called vectors; restrictions form a subspace of the parent vector space.)

Given row array $\mathbf{a}:(1\times 3)$ and column array $\mathbf{b}:(2\times 1)$ , the tensor (or outer) product $\mathbf{a}\otimes\mathbf{b}$ creates the following rectangular array:

$$\mathbf{a}\otimes\mathbf{b}:(2\times 3) := \begin{pmatrix} a_{11}\cdot\mathbf{b} & a_{12}\cdot\mathbf{b} & a_{13}\cdot\mathbf{b} \end{pmatrix} := \begin{pmatrix} a_{11}\cdot\begin{pmatrix} b_{11} \\ b_{21} \end{pmatrix} & a_{12}\cdot\begin{pmatrix} b_{11} \\ b_{21} \end{pmatrix} & a_{13}\cdot\begin{pmatrix} b_{11} \\ b_{21} \end{pmatrix} \end{pmatrix}$$
$$:= \begin{pmatrix} a_{11}\cdot b_{11} & a_{12}\cdot b_{11} & a_{13}\cdot b_{11} \\ a_{11}\cdot b_{21} & a_{12}\cdot b_{21} & a_{13}\cdot b_{21} \end{pmatrix}$$

Note that an array of one row and three columns combined with an array of two rows and one column yields an array of two rows and three columns. The tensor product of an n-dimensional row array with an m-dimensional column array is an array with m-rows and n-columns: a notational shorthand is $(1\times n):\otimes(m\times 1) := (m\times n):$. A rectangular array is also called a matrix.

In Excel, the tensor product is supported by Excel array multiplication. For example, let's first use the Define Name command (on the Formulas tab) to define the $(1\times 3)$ cell region a_ as C4:E4 and $(2\times 1)$ cell region b_ as G4:G5. Populate sample values {1,2,3} and {4,5} into these regions. Enter array formula = a_*b_ in the $(2\times 3)$ region I4:K5 (via <control-shift-enter>). Excel inserts the curly brackets to show it is an array formula ={a_*b_} (unfortunately these brackets do not show when printed). Array multiplication yields the correct answer for tensor product (see Figure 2.1).

| | A | B | C | D | E | F | G | H | I | J | K |
|---|---|---|---|---|---|---|---|---|---|---|---|
| 3 | | | a_ | | | | b_ | | a_ * b_ | | |
| 4 | | | 1 | 2 | 3 | | 4 | | 4 | 8 | 12 |
| 5 | | | | | | | 5 | | 5 | 10 | 15 |

| | A | B | C | D | E | F | G | H | I | J | K |
|---|---|---|---|---|---|---|---|---|---|---|---|
| 3 | | | a_ | | | | b_ | | a_ * b_ | | |
| 4 | | | 1 | 2 | 3 | | 4 | | =a_*b_ | =a_*b_ | =a_*b_ |
| 5 | | | | | | | 5 | | =a_*b_ | =a_*b_ | =a_*b_ |
| 6 | | | | | | | | | | | |

**Figure 2.1. Tensor product of two arrays: $=\text{a_}*\text{b_}$ corresponds to $\mathbf{a}\otimes\mathbf{b}$.**
**Array a_ is defined as C4:E4 and array b_ as G4:G5. Top: Values. Bottom: Formulas.**

Mathematical notations are more forgiving than programming languages. Let's extend the mathematical notation and introduce an Excel-like spreadsheet notation for specifying arrays. The following specifies the location of arrays a_ and b_ :

$$\text{a_} : (1\times 3)//\text{C4}:\text{E4}$$
$$\text{b_} : (2\times 1)//\text{G4}:\text{G5}$$

Given a name, define a restriction:

$$\text{a_} : (1\times 2)//\text{C4}:\text{D4}$$

Here is a notation for defining array formulas. An example: the (un-named) region I4:K5 contains the array (tensor) product of a_ and b_ :

$$:(2\times 3)//\text{I4}:\text{K5} \;:=\; \text{a_}*\text{b_}$$

Backpropagation uses other array operations. Matrix addition for two matrices having the same number of rows and columns is just array addition in Excel. Mathematically:

$$\mathbf{w} := \begin{pmatrix} w_{11} & w_{12} & w_{13} \\ w_{21} & w_{22} & w_{23} \end{pmatrix} \text{ and } \mathbf{v} := \begin{pmatrix} v_{11} & v_{12} & v_{13} \\ v_{21} & v_{22} & v_{23} \end{pmatrix} \text{ then}$$

$$\mathbf{w}+\mathbf{v} := \begin{pmatrix} w_{11}+v_{11} & w_{12}+w_{12} & w_{13}+v_{13} \\ w_{21}+v_{21} & w_{22}+v_{22} & w_{23}+v_{23} \end{pmatrix}.$$

Excel implements the Hadamard product $\mathbf{w}\odot\mathbf{v}$ for two arrays having the same number of rows and columns as array multiplication:

$$\mathbf{w}\odot\mathbf{v}=\begin{pmatrix} w_{11} & w_{12} & w_{13} \\ w_{21} & w_{22} & w_{23}\end{pmatrix}\odot\begin{pmatrix} v_{11} & v_{12} & v_{13} \\ v_{21} & v_{22} & v_{23}\end{pmatrix}:=\begin{pmatrix} w_{11}\cdot v_{11} & w_{12}\cdot v_{12} & w_{13}\cdot v_{13} \\ w_{21}\cdot v_{21} & w_{22}\cdot v_{22} & w_{23}\cdot v_{23}\end{pmatrix} \text{ and}$$

$$\mathbf{b}\odot\mathbf{b}:(2\times 1):=\begin{pmatrix} b_{11}\cdot b_{11} \\ b_{21}\cdot b_{21}\end{pmatrix}=\begin{pmatrix} b_{11}^2 \\ b_{21}^2\end{pmatrix} \quad .$$

We define matrix addition and Hadamard product for restrictions as well: for example:

$$\mathbf{w}\odot\mathbf{v}:(2\times 2):=\begin{pmatrix} w_{11} & w_{12} & w_{13} \\ w_{21} & w_{22} & w_{23}\end{pmatrix}\odot\begin{pmatrix} v_{11} & v_{12} & v_{13} \\ v_{21} & v_{22} & v_{23}\end{pmatrix}:(2\times 2):=\begin{pmatrix} w_{11}\cdot v_{11} & w_{12}\cdot v_{12} \\ w_{21}\cdot v_{21} & w_{22}\cdot v_{22}\end{pmatrix} \quad .$$

How does an Excel spreadsheet tell the difference between Hadamard product and tensor product? By context: Hadamard product `=a_*b_` is defined for two arrays having the same number of rows and columns.

Another important operation used throughout the neural network and machine learning literature is applying a real-valued function to every component in an array. Examples of real-valued functions that are built-in Excel worksheet functions are $f(x)=\exp(x)$, $f(x)=\tanh(x)$, and the Heaviside step function $f(x)=u(x):=if(x>0,1,0)$. For example, given array $\mathbf{w}:(2\times 3)$ defined above, define $f(\mathbf{w})$ as:

$$f(\mathbf{w}):(2\times 3):=f(\mathbf{w}:(2\times 3)):(2\times 3):=\begin{pmatrix} f(w_{11}) & f(w_{12}) & f(w_{13}) \\ f(w_{21}) & f(w_{22}) & f(w_{23})\end{pmatrix}$$

For $f(x)=\tanh(x)$:

$$f(\mathbf{a}^T):(3\times 1):=\tanh(\mathbf{a}^T):(3\times 1):=\begin{pmatrix} \tanh(a_{11}) \\ \tanh(a_{21}) \\ \tanh(a_{31})\end{pmatrix}$$

Applying a functions to an array is called a map in the functional programming literature: this is supported with spreadsheet array formulas. See Figure 2.2 for Excel examples. Note that a functional map is a special case of a vector-valued function, which specifies, in general, a function for each component. An example of a vector-valued function is

$$\mathbf{f}(\mathbf{a}^T):(3\times 1)=\begin{pmatrix} f_1(\mathbf{a}^T) \\ f_2(\mathbf{a}^T) \\ f_3(\mathbf{a}^T)\end{pmatrix}:=\begin{pmatrix} \exp(a_{11}\cdot a_{12}) \\ \tanh(a_{21}) \\ if(a_{31}>0,1,0)\end{pmatrix}$$

(The vector-valued function is an array so we represent it in bold type.) The functional maps of real-valued functions that are used in backpropagation are called "activation functions." Some typical activation functions (and their first derivatives) are shown in the table in Figure 2.3. See [14] for a nice table of activation functions.

| | A | B | C | D | E | F | G | H | I | J | K | L |
|---|---|---|---|---|---|---|---|---|---|---|---|---|
| 10 | | **w_** | | | | **v_** | | | | **w_+v_** | | |
| 11 | | 1 | 2 | 3 | | -5 | -4 | 0 | | -4 | -2 | 3 |
| 12 | | 4 | 5 | 6 | | -1 | 1 | 1 | | 3 | 6 | 7 |
| 13 | | | | | | | | | | | | |
| 14 | | **w_ * v_** (2x3) | | | | **w_*v_** (2x2) | | | | exp(**v_**) | | |
| 15 | | -5 | -8 | 0 | | -5 | -8 | | | 0.01 | 0 | 1 |
| 16 | | -4 | 5 | 6 | | -4 | 5 | | | 0.37 | 2.7 | 2.72 |
| 17 | | | | | | | | | | | | |

| | A | B | C | D | E | F | G | H | I | J | K | L |
|---|---|---|---|---|---|---|---|---|---|---|---|---|
| 10 | | **w_** | | | | **v_** | | | | **w_+v_** | | |
| 11 | | 1 | 2 | 3 | | -5 | -4 | 0 | | =w_+v_ | =w_+v_ | =w_+v_ |
| 12 | | 4 | 5 | 6 | | -1 | 1 | 1 | | =w_+v_ | =w_+v_ | =w_+v_ |
| 13 | | | | | | | | | | | | |
| 14 | | **w_ * v_** (2 | | | | **w_*v_** (2x | | | | exp(**v_**) | | |
| 15 | | =w_*v_ | =w_*v_ | =w_*v_ | | =w_*v_ | =w_*v_ | | | =EXP(v_) | =EXP(v_ | =EXP(v_) |
| 16 | | =w_*v_ | =w_*v_ | =w_*v_ | | =w_*v_ | =w_*v_ | | | =EXP(v_) | =EXP(v_ | =EXP(v_) |
| 17 | | | | | | | | | | | | |

**Figure 2.2. Spreadsheet showing values and formulas for array addition `w_+_v`, Hadamard product `w_*_v`, array function mapping `exp(v_)`. Array `w_` is specified with Excel Define Name command as B11:D12; array `v_` is specified with Excel Define Name command as F11:H12. Note: `=w_+v_` corresponds to $\mathbf{w}+\mathbf{v}$; `=w_*v_` corresponds to $\mathbf{w}\odot\mathbf{v}$ ; `=exp(v_)` corresponds to $\exp(\mathbf{v})$ . Top: Values. Bottom: Formulas.**

| Function: Math/Excel | Name | Derivative |
|---|---|---|
| $f(z) := \tanh(z)$<br>`=tanh(z)` | Hyperbolic tangent | $f'(z) = 1 - (f(z))^2$ |
| $f(z) := 1/(1+\exp(-z))$<br>`=1/(1+exp(z))` | Logistic sigmoid | $f'(z) = f(z) * (1 - f(z))$ |
| $f(z) := z$<br>`=z` | Identity function (ID) | $f'(z) = 1$ |
| $f(z) := z \cdot u(z)$<br>`=if(z>0,z,0)` | Rectified Linear Unit (RELU) | $f'(z) = u(z)$<br>`=if(z>0,1,0)` |

**Figure 2.3. Some activation functions.**

The Heaviside unit step function helps specify the "rectified linear unit" or RELU function. Note that that the derivative of the Heaviside function is not defined for $z = 0$.

Backpropagation uses matrix multiplication. To review: given $(2\times 3)$ array $\mathbf{w}$ and $(3\times 1)$ array $\mathbf{b}$, the matrix product $\mathbf{w}\bullet\mathbf{b}$ is the array defined by

$$\mathbf{w}\bullet\mathbf{b}=\begin{pmatrix} w_{11} & w_{12} & w_{13} \\ w_{21} & w_{22} & w_{23} \end{pmatrix}\bullet\begin{pmatrix} b_{11} \\ b_{21} \\ b_{31} \end{pmatrix}:=\begin{pmatrix} w_{11}\cdot b_{11}+w_{12}\cdot b_{21}+w_{13}\cdot b_{31} \\ w_{21}\cdot b_{11}+w_{22}\cdot b_{21}+w_{23}\cdot b_{31} \end{pmatrix}$$

Matrix product is supported in Excel by the MMULT worksheet function. In general, given $(n\times m)$ array $\mathbf{w}$ and $(m\times 1)$ array $\mathbf{b}$, the matrix product $\mathbf{w}\bullet\mathbf{b}$ is an $(n\times 1)$ array with each component given by a sum of products. A shorthand for this rule is $:(n\times m)\bullet:(m\times 1)=:(n\times 1)$.

We combine matrix multiply with the $(row\times column)$ operator for a restriction. For example:

$$\begin{pmatrix} w_{11} & w_{12} & w_{13} \\ w_{21} & w_{22} & w_{23} \\ 0 & 1 & 1 \end{pmatrix}\bullet\begin{pmatrix} b_{11} \\ b_{21} \\ 1 \end{pmatrix}:(2\times 1):=\begin{pmatrix} w_{11}\cdot b_{11}+w_{12}\cdot b_{21}+w_{13} \\ w_{21}\cdot b_{11}+w_{22}\cdot b_{21}+w_{23} \end{pmatrix}$$

In Excel, restrictions supported by selecting a partial range for the formula. See Figure 2.4.

| | A | B | C | D | E | F | G | H | I | J | K | L |
|---|---|---|---|---|---|---|---|---|---|---|---|---|
| 23 | | | | | | | | w_1 • x | | | | |
| 24 | | w_1 | | | | x | | (3x1) | | (2x1) | | (1x1) |
| 25 | | 1 | 2 | 3 | | 3 | | 16 | | 16 | | 16 |
| 26 | | 4 | 5 | 6 | | 5 | | 43 | | 43 | | |
| 27 | | 0 | 1 | 1 | | 1 | | 6 | | | | |
| 28 | | | | | | | | | | | | |

| | A | B | C | D | E | F | G | H | I | J | K | L |
|---|---|---|---|---|---|---|---|---|---|---|---|---|
| 23 | | | | | | | | w_1 • x | | | | |
| 24 | | w_1 | | | | x | | (3x1) | | (2x1) | | (1x1) |
| 25 | | 1 | 2 | 3 | | 3 | | =MMULT(w_1,x) | | =MMULT(w_1,x) | | =MMULT(w_1,x) |
| 26 | | 4 | 5 | 6 | | 5 | | =MMULT(w_1,x) | | =MMULT(w_1,x) | | |
| 27 | | 0 | 1 | 1 | | 1 | | =MMULT(w_1,x) | | | | |
| 28 | | | | | | | | | | | | |

**Figure 2.4. Matrix multiplication for arrays `w_` and `a_` (these arrays were specified in Figures 2.1 and 2.3.) Note the three different restrictions for the matrix multiplication. The "usual" result is $(3\times 1)$ ; however we also show the $(2\times 1)$ and $(1\times 1)$ restrictions. Note: `=MMult(w_1,x)` corresponds to $\mathbf{w}_1\bullet\mathbf{x}$.**

### 2.2 Generalizing Linear Regression

Neural network nonlinear formulas generalize the formulas used in linear regression. Let's first review some linear regression formulas. Given the following arrays:

$$\mathbf{x}:(n\times 1):=\begin{pmatrix} x_1 \\ \vdots \\ x_n \end{pmatrix};\ \mathbf{out}:(m\times 1):=\begin{pmatrix} out_1 \\ \vdots \\ out_m \end{pmatrix};\ \mathbf{b}:(m\times 1):=\begin{pmatrix} b_1 \\ \vdots \\ b_m \end{pmatrix};\ \ \mathbf{A}:(m\times n)$$

a linear formula that relates $\mathbf{x}$ to $\mathbf{out}$ is:

$$\mathbf{out}:(m\times 1):=\mathbf{A}\bullet\mathbf{x}+\mathbf{b}$$

Mathematicians call this linear formula an affine transformation: it transforms $(n\times 1)$ vectors to $(m\times 1)$ vectors. (This formula generalizes the equation for a line: $y=a\cdot x+b$ . )

Here is an alternative representation. Recall that the result of the matrix multiply $\mathbf{A}\bullet\mathbf{x}$ is a column array via rule: $(m\times n)\bullet:(n\times 1)=:(m\times 1)$ . Extend the $m\times n$ array $\mathbf{A}$ to rectangular array $\mathbf{w}$ with an extra column whose values are the column vector $\mathbf{b}$:

$$\mathbf{w}:(m\times(n+1)):=\begin{pmatrix}\mathbf{A} & \mathbf{b}\end{pmatrix}:(m\times(n+1))\ \text{with}\ w_{i,m+1}=b_i,\ i=1,2,...,n.$$

Array $\mathbf{w}$ is sometimes called a weight matrix. Now define column vector $\mathbf{inp}$ as an extension of $\mathbf{x}$:

$$\mathbf{inp}:((n+1)\times 1):=\begin{pmatrix}\mathbf{x} \\ 1\end{pmatrix}:=\begin{pmatrix} x_1 \\ \vdots \\ x_n \\ 1 \end{pmatrix}\ \text{with}\ inp_{n+1}=1.$$

Then an alternative representation of a linear formula is:

$$\mathbf{out}:=\mathbf{w}\bullet\mathbf{inp}:=\begin{pmatrix}\mathbf{A} & \mathbf{b}\end{pmatrix}\bullet\begin{pmatrix}\mathbf{x} \\ 1\end{pmatrix}:=\mathbf{A}\bullet\mathbf{x}+\mathbf{b}$$

In this representation, a linear formula that relates $\mathbf{inp}$ to $\mathbf{out}$ is just the matrix multiply of $\mathbf{w}$ with $\mathbf{inp}$: we save a matrix addition operation by subsuming $\mathbf{b}$ in $\mathbf{w}$. Next, suppose we are given a set of array pairs:

$$(\mathbf{x}(1),\mathbf{targ}(1)),(\mathbf{x}(2),\mathbf{targ}(2)),\ldots,(\mathbf{x}(s),\mathbf{targ}(s))\ldots,(\mathbf{x}(S),\mathbf{targ}(S))$$

Or equivalently,

$$(\mathbf{inp}(1),\mathbf{targ}(1)),\ldots,(\mathbf{inp}(S),\mathbf{targ}(S))$$

with

$$\mathbf{inp}(s):((n+1)\times 1):=\begin{pmatrix}\mathbf{x}(s)\\ 1\end{pmatrix}:=\begin{pmatrix}x_1(s)\\ \vdots\\ x_n(s)\\ 1\end{pmatrix} \text{ and } \mathbf{targ}(s):(m\times 1):=\begin{pmatrix}targ_1(s)\\ \vdots\\ targ_m(s)\end{pmatrix}$$

Each column array pair is a called a sample record or a training record. For some arbitrary $m\times(n+1)$ array $\mathbf{w}$, compute the output as a linear formula:

$$\mathbf{out}(s):(m\times 1):=\begin{pmatrix}out_1(s)\\ \vdots\\ out_m(s)\end{pmatrix}:=\mathbf{w}\bullet\mathbf{inp}(s) \text{ for } s=1,2,\cdots,S \ .$$

The linear regression problem is: Finds a linear formula (array $\mathbf{w}$) that – given an input vector $\mathbf{inp}$ – computes a vector $\mathbf{out}$ that is optimal in the following sense: if the input vector is one of the sample input vectors $\mathbf{inp}(s)$, then the output vector $\mathbf{out}$ should be "approximately equal to" the corresponding target vector $\mathbf{targ}(s)$. In other words:

$$\mathbf{out}(s):=\mathbf{w}\bullet\mathbf{inp}(s)\approx\mathbf{targ}(s)$$

The error for each sample record is

$$\mathbf{targ}(s)-\mathbf{out}(s)$$

For a perfect approximation, the error for each sample should be zero. In practice this is rarely the case due to improper observation, "noise," or computational faults. The same inputs could even produce different targets at different times: this can be due to accounting errors or due to a misspecification that mistakenly neglected "hidden" inputs. In any case, a common criteria is find the $\mathbf{w}_{\text{opt}}$ that minimizes an average error over all samples. Because of this, the output might not be an exact match: it could correspond to an average of the corresponding targets.

Here are three methods that find $\mathbf{w}_{\text{opt}}$:

**Method I.** Randomly guess many values for every component of array $\mathbf{w}$ – denote each random guess by $\tilde{\mathbf{w}}$ – and with this guess matrix, compute the average errors over the entire sample. One average error formula computes the mean square error over the sample:

$$\text{err}:=\frac{1}{S}\cdot\sum_{s=1}^{S}\left(\mathbf{targ}(s)-\mathbf{out}(s;\tilde{\mathbf{w}})\right)^2 \text{ for each guess } \tilde{\mathbf{w}}\,.$$

If we find this error "good enough" then stop the procedure and use the matrix that had the smallest average error as "the solution" for $\mathbf{w}_{\text{opt}}$ ; otherwise continue guessing (viz, in machine learning:

"training"). It can be shown using results from probability (Monte Carlo theory), that if we keep guessing long enough (and the initial guess is not an unlucky one), we will closely approximate the optimal $\mathbf{w}_{\text{opt}}$ that minimizes err *over all possible guesses*. (Mathematically: the weights converge or are "trained" to approximate the optimum solution.) Method I is iterative: it requires many repetitions (iterations) of the same steps in order to achieve an answer. Sometimes the process of iteration is called "training" or "learning."

**Method II.** Guess just one starting value for every component of array $\mathbf{w}$. Next, we pick an arbitrary record in the sample (either in order, shuffled order, or completely at random). Call this sample record number $r$. Using the current array $\mathbf{w}$ and the selected record $r$, use $\mathbf{inp}(r)$ to compute $\mathbf{out}(r)$ and then compute the error of sample number $r$: $\mathbf{targ}(r)-\mathbf{out}(r)$. Maintain a running average of errors over the entire sample. If we find the running average of the errors "good enough," then stop. Otherwise, perform a tensor product of this error with the transpose of the input. (Recall : $(1\times(n+1)):\otimes(m\times 1):=(m\times(n+1)):$ ). Multiply this by a small positive number (traditionally called $\eta$ – eta – usually less than one). Add this to the current array $\mathbf{w}$ to create a new updated array. The result is the new weight matrix:

$$\mathbf{w}_{\text{new}} := \mathbf{w} + \eta\cdot\left(\mathbf{inp}(s)\right)^T \otimes \left(\mathbf{targ}(s)-\mathbf{out}(s)\right) \;;$$

Repeat the process: set $\mathbf{w}:=\mathbf{w}_{\text{new}}$ and pick another sample. Recompute the error with this current array $\mathbf{w}$. If we the running average of the errors is "good enough," then stop. Otherwise continue the iteration. This method, due to Widrow [15] (dating from 1959) is sometimes called the delta rule. It can be shown (using differential calculus) that the updated iterates will closely approximate the optimal $\mathbf{w}_{\text{opt}}$ that minimizes err over all possible guesses – provided that we wait long enough and the initial guess is not an unlucky one. Method II is also iterative: at each iteration, we are mathematically taking a small step along the gradient – the direction of steepest decrease of the errors (visualize the error surface as a bowl). Backpropagation is a generalization of Method II.

**Method III.** This method (called the classical method of least squares) is at least 200 years old. It is the basis of Excel's LINEST and TREND formulas. The first step is to consolidate all inputs and targets in their own arrays:

$$\mathbf{INPUTS}:((n+1)\times S):=\left(\mathbf{inp}(1):(n+1)\times 1 \quad \mathbf{inp}(2):(n+1)\times 1 \quad \ldots \quad \mathbf{inp}(S):(n+1)\times 1\right)$$
$$\mathbf{TARGETS}:(m\times S):=\left(\mathbf{targ}(1):m\times 1 \quad \mathbf{targ}(2):m\times 1 \quad \ldots \quad \mathbf{targ}(S):m\times 1\right)$$

Array $\mathbf{INPUTS}$ has, as columns, all input samples $\mathbf{inp}(s):(n+1)\times 1$; array $\mathbf{TARGETS}$ has, as columns, all corresponding target samples $\mathbf{targ}(s):m\times 1$. It can be shown (using differential calculus), the optimal $\mathbf{w}_{\text{opt}}$ that minimizes err over all possible guesses is:

$$\mathbf{w}_{\text{opt}} : (m \times (n+1)) := \Big[ \big( \mathbf{INPUTS} : ((n+1) \times S) \bullet \mathbf{INPUTS}^T : (S \times (n+1)) \big)^{-1}$$
$$\bullet \big( \mathbf{INPUTS} : ((n+1) \times S) \bullet \mathbf{TARGETS}^T : (S \times m) \big) \Big]^T$$

This solution uses the matrix inverse operation: for matrix $\mathbf{A}$ its inverse is denoted by $\mathbf{A}^{-1}$ . Matrix inverse is a worksheet function supported by Excel. The copy-and-paste array formula for the regression weights is

```
=TRANSPOSE (   MMULT( MINVERSE( MMULT(INPUTS,TRANSPOSE(INPUTS))),
               MMULT(INPUTS,TRANSPOSE(TARGETS) )  )  )
```

The advantage of Method III is that it is not an iterative approximation: it is exact (as long as the data is "well-conditioned" in the numerical analysis sense). Figure 2.5 shows a simple spreadsheet implementation of a regression problem.

| | A | B | C | D | E | F |
|---|---|---|---|---|---|---|
| 1 | | | | | | |
| 2 | | **Samples** | *x1* | *x2* | *targ1* | *targ2* |
| 3 | | *0* | 0 | 0 | 0 | 0 |
| 4 | | *1* | 0 | 1 | 1 | 0 |
| 5 | | *2* | 1 | 0 | 1 | 0 |
| 6 | | *3* | 1 | 1 | 0 | 1 |
| 7 | | | | | | |
| 8 | | INPUTS | *inp(0)* | *inp(1)* | *inp(2)* | *inp(3)* |
| 9 | | *x1* | 0 | 0 | 1 | 1 |
| 10 | | *x2* | 0 | 1 | 0 | 1 |
| 11 | | | 1 | 1 | 1 | 1 |
| 12 | | | | | | |
| 13 | | TARGETS | *targ(0)* | *targ(1)* | *targ(2)* | *targ(3)* |
| 14 | | *targ1* | 0 | 1 | 1 | 0 |
| 15 | | *targ2* | 0 | 0 | 0 | 1 |
| 16 | | | | | | |
| 17 | | wopt | | | | |
| 18 | | | 1.11E-16 | 1.11E-16 | 0.5 | |
| 19 | | | 0.5 | 0.5 | -0.25 | |
| 20 | | | | | | |

| **Samples** | *x1* | *x2* | *targ1* | *targ2* | ***out1*** | ***out2*** |
|---|---|---|---|---|---|---|
| *0* | 0 | 0 | 0 | 0 | 0.5 | -0.25 |
| *1* | 0 | 1 | 1 | 0 | 0.5 | 0.25 |
| *2* | 1 | 0 | 1 | 0 | 0.5 | 0.25 |
| *3* | 1 | 1 | 0 | 1 | 0.5 | 0.75 |

**Figure 2.5. Simple regression problem: create a formula from (x1,x2) that computes for output (targ1, targ2). Left: Problem specification: inputs and target values (Samples); INPUTS array; TARGETS array; optimum linear regression weights via Method III. Right: computed outputs. The formula for the optimal weights in cells C18:E19 is**

```
=TRANSPOSE(     MMULT(MINVERSE ( MMULT( C9:F11,TRANSPOSE( C9:F11))),
                MMULT(C9:F11,TRANSPOSE(C14:F15) )  )  )  .
```

Note that the sum of the squared errors are 1 (for targ1) and 0.25 (for targ2). That these errors are not zero (or even small) is due to the nature of the problem. The first target is the "exclusive or" function (returns 1 if only one input is 1 but not both; otherwise return 0); the second is the "and" function (returns 1 if both inputs are 1; otherwise return 0). We can show that no linear formula exists that exactly computes these two functions (they are not "linearly separable").

Can we make the regression more powerful in that we can make better approximations than the linear one? One way is to simply include more inputs (and arrange these inputs to be functions of other inputs). This is the functional link approach advocated by Pao [4]. Another way is to consider making the linear formula into a network of linear formulas: here, the output of one linear formula is the input to a second linear formula. Let's see how this works for a network with the first output feeding the second output and the second output feeding the third output (a "three-layered" network):

$$\begin{aligned}
&\mathbf{out}_1 : (m_1 \times 1) := \mathbf{A}_1 \bullet \mathbf{x} + \mathbf{b}_1; && \mathbf{A}_1 : (m_1 \times n); \quad \mathbf{b}_1 : (m_1 \times 1)\\
&\mathbf{out}_2 : (m_2 \times 1) := \mathbf{A}_2 \bullet \mathbf{out}_1 + \mathbf{b}_2; && \mathbf{A}_2 : (m_2 \times m_1); \quad \mathbf{b}_2 : (m_2 \times 1)\\
&\mathbf{out} := \mathbf{out}_3 : (m \times 1) := \mathbf{A}_3 \bullet \mathbf{out}_2 + \mathbf{b}_3; && \mathbf{A}_3 : (m \times m_2); \quad \mathbf{b}_3 : (m \times 1)
\end{aligned}$$

It looks like we have complicated things. But we really have not. Since

$$\begin{aligned}
\mathbf{out} &:= \mathbf{A}_3 \bullet \mathbf{out}_2 + \mathbf{b}_3\\
&:= \mathbf{A}_3 \bullet \left(\mathbf{A}_2 \bullet \mathbf{out}_1 + \mathbf{b}_2\right) + \mathbf{b}_3\\
&:= \mathbf{A}_3 \bullet \left(\mathbf{A}_2 \bullet \left(\mathbf{A}_1 \bullet \mathbf{x} + \mathbf{b}_1\right) + \mathbf{b}_2\right) + \mathbf{b}_3\\
&:= \mathbf{A}_3 \bullet \left(\mathbf{A}_2 \bullet \mathbf{A}_1 \bullet \mathbf{x} + \mathbf{A}_2 \bullet \mathbf{b}_1 + \mathbf{b}_2\right) + \mathbf{b}_3\\
&:= \left(\mathbf{A}_3 \bullet \mathbf{A}_2 \bullet \mathbf{A}_1\right) \bullet \mathbf{x} + \left(\mathbf{A}_2 \bullet \mathbf{b}_1 + \mathbf{A}_3 \bullet \mathbf{A}_2 \bullet \mathbf{b}_2 + \mathbf{b}_3\right)
\end{aligned}$$

This implies that this three-layer linear network can be reduced to a single-layer standard linear regression. The single layer has matrix $\mathbf{A}$ and column vector $\mathbf{b}$ given by:

$$\mathbf{A} := \left(\mathbf{A}_3 \bullet \mathbf{A}_2 \bullet \mathbf{A}_1\right) \text{ and } \mathbf{b} := \left(\mathbf{A}_2 \bullet \mathbf{b}_1 + \mathbf{A}_3 \bullet \mathbf{A}_2 \bullet \mathbf{b}_2 + \mathbf{b}_3\right)$$

Linear formulas of linear formulas are linear formulas. However, consider a simple extension of the linear world to the nonlinear world. Select any three activation functions (see Figure 2.3) as a map. Such a three-layered (nonlinear) network is specified by:

$$\begin{aligned}
&\mathbf{out}_1 : (m_1 \times 1) := f_1\left(\mathbf{A}_1 \bullet \mathbf{x} + \mathbf{b}_1\right)\\
&\mathbf{out}_2 : (m_2 \times 1) := f_2\left(\mathbf{A}_2 \bullet \mathbf{out}_1 + \mathbf{b}_2\right)\\
&\mathbf{out} = \mathbf{out}_3 : (m \times 1) := f_3\left(\mathbf{A}_3 \bullet \mathbf{out}_2 + \mathbf{b}_3\right)
\end{aligned}$$

Here it again looks like we have complicated things. And we really have, since nonlinear functions of nonlinear functions do not in general reduce to known simpler forms:

$$\begin{aligned}\mathbf{out} &:= f_3\left(\mathbf{A}_3 \bullet \mathbf{out}_2 + \mathbf{b}_3\right)\\ &:= f_3\left(\mathbf{A}_3 \bullet \left( f_2\left(\mathbf{A}_2 \bullet \mathbf{out}_1 + \mathbf{b}_2\right)\right) + \mathbf{b}_3\right)\\ &:= f_3\left(\mathbf{A}_3 \bullet \left( f_2\left(\mathbf{A}_2 \bullet \left( f_1\left(\mathbf{A}_1 \bullet \mathbf{x} + \mathbf{b}_1\right)\right) + \mathbf{b}_2\right)\right) + \mathbf{b}_3\right)\end{aligned}$$

For linear networks additional complexity makes no difference. Nonlinearity is where complexity creates an emergent property: nonlinear neural network regression frequently outperforms linear regression: the average errors for nonlinear neural networks are smaller than the linear regression case. They can also solve problems that are not linearly separable.

### 2.3 Nonlinear Formulas for Feed-Forward Networks

Specify an integer $q$ (usually called the number of layers, where $q = 2, 3, \ldots$): we need to account for $q$ different weight matrices and $q$ different output column vectors. As in the linear case, the training data set consists of a set of records of input-outputs. Suppose we are given a weight matrix $\mathbf{w}_1 : (m_1 \times (n+1))$; we compute the first output $\mathbf{out}_1 : ((m_1+1)\times 1)$ from the inputs, weights, and activation function $f_1$. The specifications are:

$$\mathbf{inp} : ((n+1)\times 1) := \begin{pmatrix} \mathbf{x} : (n \times 1) \\ 1 \end{pmatrix} = \begin{pmatrix} x_1 \\ \vdots \\ x_n \\ 1 \end{pmatrix};$$

$$\mathbf{w}_1 : (m_1 \times (n+1)); \qquad \mathbf{out}_1 : ((m_1+1)\times 1) := \begin{pmatrix} f_1(\mathbf{w}_1 \bullet \mathbf{inp}) : (m_1 \times 1) \\ 1 \end{pmatrix}$$

Vector $\mathbf{out}_1$ has $(m_1+1)$ rows; the restriction is $\mathbf{out}_1 : (m_1 \times 1) = f_1(\mathbf{w}_1 \bullet \mathbf{inp}) : (m_1 \times 1)$. Array $\mathbf{out}_1$ specifies a first layer. Continue the accounting by computing a second output from the first output, with a second weight matrix $\mathbf{w}_2 : (m_2 \times (m_1+1))$ and second activation function $f_2$ :

$$\mathbf{w}_2 : m_2 \times (m_1+1); \qquad \mathbf{out}_2 : ((m_2+1)\times 1) := \begin{pmatrix} f_2(\mathbf{w}_2 \bullet \mathbf{out}_1) : (m_2 \times 1) \\ 1 \end{pmatrix}$$

In general

$$\mathbf{w}_h : (m_h \times (m_{h-1}+1)); \;\; \mathbf{out}_h : ((m_h+1)\times 1) = \begin{pmatrix} f_h(\mathbf{w}_h \bullet \mathbf{out}_{h-1}) : (m_h \times 1) \\ 1 \end{pmatrix}$$

(We can set $\mathbf{out}_0 := \mathbf{inp}$.) Form the final output vector $\mathbf{out} : (m_q \times 1)$ where $m_q := m$ :

$$\mathbf{out} = \mathbf{out}_q : (m_q \times 1) = f_q(\mathbf{w}_q \bullet \mathbf{out}_{q-1}) : (m_q \times 1)$$

Note the final output is not extended to an extra dimension. We specify this computational topology via dash notation: $n - m_1 - m_2 \cdots - m_h \ldots - m_{q-1} - m_q$. Here, $n$ is the number of sample inputs and $m$ is the number of outputs. For $q \geq 2$ then the $m_h$ values refer to hidden layers: $m_1$ refers to the output dimension ("number of neurons") in the first hidden layer, $m_h$ is the number of neurons in hidden layer *h*, and so on. The final output vector $\mathbf{out}$ depends on $\mathbf{out}_{q-1}, \mathbf{out}_{q-2}, \ldots, \mathbf{out}_1$. All these must be previously computed in order of dependency. This arrangement specifies a "feed-forward network." Consequently, given an input vector and a sequence of weight matrices, we compute the output by forward-propagating the subsequent outputs as inputs to the next layer. Knowledge of input $\mathbf{inp}$, weights $\mathbf{w}_h$, and functions $f_h$ are sufficient to determine $\mathbf{out}$.

Figure 2.6 shows a 2-3-2 network in our representation. We can visualize the network topology by using the Excel Trace Dependents utility. Any change in the values in cells B5:B7 propagates to the final outputs in cells K5:K6. The array specifications are:

$$\mathbf{inp} : ((2+1)\times 1) := \begin{pmatrix} 1 \\ 0 \\ 1 \end{pmatrix} //B5:B7$$

$$\mathbf{w}_1 : (3\times(2+1)) //C5:E7; \quad \mathbf{out}_1 : ((3+1)\times 1) := \begin{pmatrix} \tanh(\mathbf{w}_1 \bullet \mathbf{inp}) : (3\times 1) \\ 1 \end{pmatrix} //F5:F8$$

$$\mathbf{w}_2 : (2\times(3+1)) //G5:J6; \quad \mathbf{out} = \mathbf{out}_2 : (2\times 1) := \tanh(\mathbf{w}_2 \bullet \mathbf{out}_1) : (2\times 1) //K5:K6$$

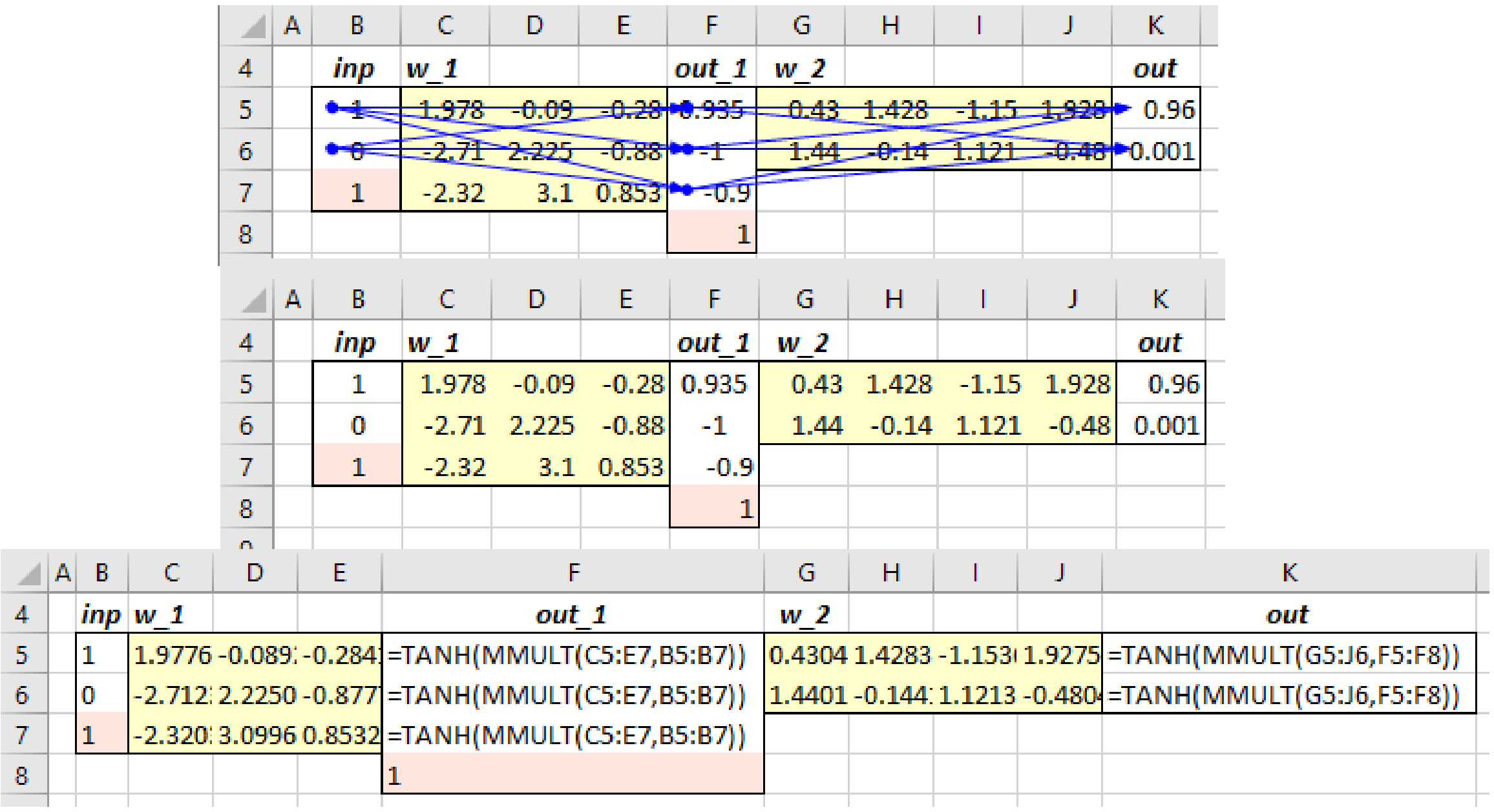

| | A | B | C | D | E | F | G | H | I | J | K |
|---|---|---|---|---|---|---|---|---|---|---|---|
| 4 | | *inp* | *w_1* | | | *out_1* | *w_2* | | | | *out* |
| 5 | | 1 | 1.978 | -0.09 | -0.28 | 0.935 | 0.43 | 1.428 | -1.15 | 1.928 | 0.96 |
| 6 | | 0 | -2.71 | 2.225 | -0.88 | -1 | 1.44 | -0.14 | 1.121 | -0.48 | 0.001 |
| 7 | | 1 | -2.32 | 3.1 | 0.853 | -0.9 | | | | | |
| 8 | | | | | | 1 | | | | | |

| | A | B | C | D | E | F | G | H | I | J | K |
|---|---|---|---|---|---|---|---|---|---|---|---|
| 4 | | *inp* | *w_1* | | | *out_1* | *w_2* | | | | *out* |
| 5 | | 1 | 1.978 | -0.09 | -0.28 | 0.935 | 0.43 | 1.428 | -1.15 | 1.928 | 0.96 |
| 6 | | 0 | -2.71 | 2.225 | -0.88 | -1 | 1.44 | -0.14 | 1.121 | -0.48 | 0.001 |
| 7 | | 1 | -2.32 | 3.1 | 0.853 | -0.9 | | | | | |
| 8 | | | | | | 1 | | | | | |

| | A | B | C | D | E | F | G | H | I | J | K |
|---|---|---|---|---|---|---|---|---|---|---|---|
| 4 | | *inp* | *w_1* | | | *out_1* | *w_2* | | | | *out* |
| 5 | | 1 | 1.9776 | -0.089 | -0.284 | =TANH(MMULT(C5:E7,B5:B7)) | 0.4304 | 1.4283 | -1.153 | 1.9275 | =TANH(MMULT(G5:J6,F5:F8)) |
| 6 | | 0 | -2.712 | 2.2250 | -0.877 | =TANH(MMULT(C5:E7,B5:B7)) | 1.4401 | -0.144 | 1.1213 | -0.480 | =TANH(MMULT(G5:J6,F5:F8)) |
| 7 | | 1 | -2.320 | 3.0996 | 0.8532 | =TANH(MMULT(C5:E7,B5:B7)) | | | | | |
| 8 | | | | | | 1 | | | | | |

**Figure 2.6. A 2-3-2 feedforward topology for that solves the regression problem specified in Figure 2.5. Top: Worksheet values view. Middle: Recreation of the neural network view via Excel's Trace Dependents utility. Bottom: Formula view.**

## 2.4 Formulas for Backpropagation

For linear formulas, the squared error surface for the weights involves squaring the linear outputs. The resultant error surface is a quadratic formula of the outputs: it corresponds to a mult-dimensional bowl. We pick a point and take iterative steps in the direction of the steepest downward slope. Slopes of quadratics are easy to compute: that is how Method II works in finding optimal weights.

For nonlinear formulas based on the functional mapping of activation functions, we need a representation of both $f(z)$ and its first derivative. (Recall the first derivative of a function $f'(z)$ corresponds to the slope of the tangent of the function at the point $z$). Backpropagation requires first derivative of our activation functions to implement gradient descent. At layer *h*, the derivative function is evaluated at column array $\mathbf{z} = \mathbf{w}_h \bullet \mathbf{out}_{h-1}$:

$$\left( f_h'(\mathbf{z})\big|_{\mathbf{z}=\mathbf{w}_h \bullet \mathbf{out}_{h-1}} \right) : (m_h \times 1) \ .$$

Figure 2.7 shows array expressions for these derivatives applied to to array $\mathbf{z} = \mathbf{w}_h \bullet \mathbf{out}_{h-1}$. Note that the result uses Hadamard product $\odot$. Even though this looks mathematically very complicated, in a spreadsheet this reduces to array multiplication.

| Function: Math/Excel | Name<br>$f_h(\mathbf{w}_h \bullet \mathbf{out}_{h-1}) = \mathbf{out}_h : (m_h \times 1)$ | Derivative<br>$\left( f_h'(\mathbf{z})\big|_{\mathbf{z}=\mathbf{w}_h \bullet \mathbf{out}_{h-1}} : (m_h \times 1) \right)$ |
|---|---|---|
| $f(z) = \tanh(z)$<br>`=tanh(z)` | Hyperbolic tangent<br>$\tanh(\mathbf{w}_h \bullet \mathbf{out}_{h-1}) = \mathbf{out}_h : (m_h \times 1)$ | $(\mathbf{1} - \mathbf{out}_h \odot \mathbf{out}_h) : (m_h \times 1)$ |
| $f(z) = 1/(1+\exp(-z))$<br>`=1/(1+exp(z))` | Logistic sigmoid<br>$1/(1+\exp(-\mathbf{w}_h \bullet \mathbf{out}_{h-1})) = \mathbf{out}_h : (m_h \times 1)$ | $(\mathbf{out}_h) \odot (\mathbf{1} - \mathbf{out}_h) : (m_h \times 1)$ |
| $f(z) = z$<br>`=z` | Identity function<br>$\mathbf{w}_h \bullet \mathbf{out}_{h-1} = \mathbf{out}_h : (m_h \times 1)$ | $\mathbf{1} : (m_h \times 1)$ |
| $f(z) = z \cdot u(z)$<br>`=if(z>0,z,0)` | Rectified Linear Unit (RELU)<br>$(\mathbf{w}_h \bullet \mathbf{out}_{h-1}) \odot u(\mathbf{w}_h \bullet \mathbf{out}_{h-1}) : (m_h \times 1)$ | $u(\mathbf{out}_h) : (m_h \times 1)$ |

**Figure 2.7. Derivatives of the activation functions listed in Figure 2.3 evaluated at column array $\mathbf{z} = \mathbf{w}_h \bullet \mathbf{out}_{h-1}$. Note that $\mathbf{1}$ denotes the h-dimensional unity column vector with all columns assigned to 1.**

The identity function $f(z) = z$ is actually the activation function for the linear formula used in Method II. As we observed, a multiple-layer linear network does not improve matters over linear regression. On the other hand, the other nonlinear functions listed in Figure 2.3 and 2.7 induce an error surface that generally is not a simple bowl: in general it is a bowl with humps and valleys. This means that a steepest descent gradient search on the error surface is not guaranteed to find an absolute smallest error. No matter: we stop when the error is "good enough"; otherwise restart with another guess.

In backpropagation, we compute a series of column arrays $\boldsymbol{\delta}$ (called deltas) that are related to activation function error gradients (slopes). For $q$ layers we need to account for $q$ deltas, one delta per output (or what is the same: one delta per weight matrix).

We proceed backwards, from the final output layer. Define the delta as the array difference $\mathbf{targ}-\mathbf{out}$ (the "error") multiplied (via Hadamard product) by the derivative (slope) of the activation function. The result is:

$$\boldsymbol{\delta}:(m_q\times 1):=\boldsymbol{\delta}_q:(m_q\times 1):=(\mathbf{targ}-\mathbf{out})\odot\left(f_q{}'(\mathbf{z})\Big|_{\mathbf{z}=\mathbf{w}_q\bullet\mathbf{out}_{q-1}}\right):(m_q\times 1):$$

For other layers, $\boldsymbol{\delta}_h$ is computed from the previously computed $\boldsymbol{\delta}_{h+1}$: errors computed previously are "back-propagated" from "outer" layers to "inner" layers. So for $h=q-1,q-2,\cdots,2,1$:

$$\boldsymbol{\delta}_h:(m_h\times 1):=\left((\mathbf{w}_{h+1})^T\bullet\boldsymbol{\delta}_{h+1}:(m_h\times 1)\right)\odot\left(f_h{}'(\mathbf{z})\Big|_{\mathbf{z}=\mathbf{w}_h\bullet\mathbf{out}_{h-1}}\right):(m_h\times 1);\ h=q-1,\cdots,1$$

Note that the matrix multiplication of the transpose of the weight matrix involves a restriction. The weight $(\mathbf{w}_{h+1}:(m_{h+1}\times(m_h+1)))^T:((m_h+1)\times m_{h+1})$ matrix multiplied by delta $\boldsymbol{\delta}_{h+1}:(m_{h+1}\times 1)$ should yield a column vector specified by $((m_h+1)\times 1)$; we take the restriction $(m_h\times 1)$. For example: consider the tanh activation function. At final output layer:

$$\boldsymbol{\delta}:(m_q\times 1):=(\mathbf{targ}-\mathbf{out})\odot(\mathbf{1}-\mathbf{out}_h\odot\mathbf{out}_h):(m_q\times 1):$$

For $h=q-1,q-2,\cdots,2,1$

$$\boldsymbol{\delta}_h:(m_h\times 1):=\left((\mathbf{w}_{h+1})^T\bullet\boldsymbol{\delta}_{h+1}:(m_h\times 1)\right)\odot(\mathbf{1}-\mathbf{out}_h\odot\mathbf{out}_h):(m_h\times 1)\quad .$$

Figure 2.8 shows the target and last two layers for a 2-2-2-2 network using tanh. Here $q=3$; $n=2;\ m_1=2;\ m_2=2;\ m_3=2$.

| targ |
|---|
| 0 |
| 1 |

| out_2 | w_3 | | | out |
|---|---|---|---|---|
| 0.766287 | -1.40842 | 1.394039 | -0.3026 | 0.006799 |
| 0.996141 | 1.388176 | -0.24171 | 1.184837 | 0.964575 |
| 1 | | | | |

| del | del_2 |
|---|---|
| -0.0068 | 0.005366 |
| 0.002465 | -7.8E-05 |

| del | del_2 |
|---|---|
| =(targ-out)*(1-out^2) | =MMULT(TRANSPOSE(w_3),del)*(1-out_2^2) |
| =(targ-out)*(1-out^2) | =MMULT(TRANSPOSE(w_3),del)*(1-out_2^2) |

**Figure 2.8. Computation of deltas $\boldsymbol{\delta}$ and $\boldsymbol{\delta}_2$. Top: values; bottom: formulas. Note:**

$$\boldsymbol{\delta}:=(\mathbf{targ}-\mathbf{out})\odot(\mathbf{1}-\mathbf{out}\odot\mathbf{out});\ \ \boldsymbol{\delta}_2:=\left((\mathbf{w}_3)^T\bullet\boldsymbol{\delta}\right)\odot((\mathbf{1}-\mathbf{out}_2\odot\mathbf{out}_2)).$$

We use the deltas to update the weights iteratively. As in Method II, let $\eta$ (eta) be a small real number: usually $0 < \eta < 1$. We need to account for a set of $q$ such etas: $\eta_1, ..., \eta_h, \ldots, \eta_q$, one for each weight matrix. Given a set of $q$ weight matrices $\mathbf{w}_1, ...\mathbf{w}_h, ..., \mathbf{w}_q$ that already have values (they all might initially have random values). Then given an input sample, we compute the output $\mathbf{out}$ (and all intermediary outputs) and compare it to the target $\mathbf{targ}$. We use the errors between target and output to update the weights using the tensor product $\otimes$ of the current delta and the previous output (which is actually the input to the current output). The backpropagation formula is:

$$\mathbf{w}_q := \mathbf{w}_q + \eta_q \cdot \left(\mathbf{out}_{q-1}\right)^T \otimes \boldsymbol{\delta}_q$$
$$\vdots$$
$$\mathbf{w}_h := \mathbf{w}_h + \eta_h \cdot \left(\mathbf{out}_{h-1}\right)^T \otimes \boldsymbol{\delta}_h ; h \le q$$
$$\vdots$$
$$\mathbf{w}_1 := \mathbf{w}_1 + \eta_1 \cdot \left(\mathbf{inp}\right)^T \otimes \boldsymbol{\delta}_1$$

Note that the arrays are dimensionally consistent: we have specifications

$$\left(\mathbf{out}_{h-1} : ((m_{h-1}+1)\times 1)\right)^T : 1\times(m_{h-1}+1) \text{ and } \boldsymbol{\delta}_h : (m_h \times 1)$$

As we showed in the previously, the tensor product of an n-dimensional row vector with an m-dimensional column vector is a matrix with m-rows and n-columns: in our case we have

$$(1\times(m_{h-1}+1) : _ \otimes (m_h \times 1) : _ = (m_h \times m_{h-1}+1) : _ \; .$$

This matrix has the same rows and columns as $\mathbf{w}^h : (m_h \times (m_{h-1}+1))$. Figure 2.9 shows the array weights in last two layers for the 2-2-2-2 network.

| | | |
|---|---|---|
| **w_3** | | |
| =w_3+eta*(TRANSPOSE(out_2)*del) | =w_3+eta*(TRANSPOSE(out_2)*del) | =w_3+eta*(TRANSPOSE(out_2)*del) |
| =w_3+eta*(TRANSPOSE(out_2)*del) | =w_3+eta*(TRANSPOSE(out_2)*del) | =w_3+eta*(TRANSPOSE(out_2)*del) |
| | | |
| **w_2** | | |
| =w_2+eta*(TRANSPOSE(out_1)*del_2) | =w_2+eta*(TRANSPOSE(out_1)*del_2) | =w_2+eta*(TRANSPOSE(out_1)*del_2) |
| =w_2+eta*(TRANSPOSE(out_1)*del_2) | =w_2+eta*(TRANSPOSE(out_1)*del_2) | =w_2+eta*(TRANSPOSE(out_1)*del_2) |

**Figure 2.9. Backpropagation updates.**
**Top: Formula View for $\mathbf{w}_3$. Bottom: Formula View for $\mathbf{w}_2$**

Note: we can use the same $\eta$ for all layers, so that $\eta = \eta_1 = ... = \eta_h = \ldots = \eta_q$. Or we can use an individual $\eta$ for each cell based on Hadamard multiplication. In this case $\boldsymbol{\eta}$ becomes a matrix. For layer *h*:

$$\boldsymbol{\eta}_h : (m_h \times (m_{h-1} + 1))$$

Learning becomes

$$\begin{aligned}
&\mathbf{w}_q := \mathbf{w}_q + \boldsymbol{\eta}_q \odot \left(\mathbf{out}_{q-1}\right)^T \otimes \boldsymbol{\delta}_q \\
&\vdots \\
&\mathbf{w}_h := \mathbf{w}_h + \boldsymbol{\eta}_h \odot \left(\mathbf{out}_{h-1}\right)^T \otimes \boldsymbol{\delta}_h ; h \le q \\
&\vdots \\
&\mathbf{w}_1 := \mathbf{w}_1 + \boldsymbol{\eta}_1 \odot \left(\mathbf{inp}\right)^T \otimes \boldsymbol{\delta}_1
\end{aligned}$$

Note that the formulas for the outputs, deltas, and weights are a functional specification for the backpropagation algorithm: formulas are presented declaratively: there are no procedural programming structures. However: before rushing to copy these formulas into an Excel spreadsheet we must note that *these formulas will not work*! We did not specify how the inputs are selected for the algorithm; nor did we specify the initial states of the weight arrays. We are close, but not close enough for a working implementation. In order to do this we need to review spreadsheet computational semantics.

## 3 Visualizing Visual Back Propagation

The backpropagation formulas are formulas are recursive ("circular" in spreadsheet terminology). To resolve circular formulas we require an understanding of the Manual Mode of Calculation. Essentially, spreadsheet computation in Manual Mode is very similar to the computation of a systolic array: a parallel architecture used for signal processing:

> A systolic array architecture is a subset of a data-flow architecture. It is made up of a number of identical cells, each cell locally connected to its nearest neighbor. The cells are usually arranged in a definite geometric pattern, corresponding to the tessellations of the Euclidean plane. The cells that are on the boundary of the pattern are able to interact with the outside world, At a given clock pulse, data enters and exits each cell; entering data is processed and stored so it can be output at the next pulse. Computational power is thus identified with the speed of input and output: a wavefront of computation is propagated in the array with a throughput proportional to the input/output bandwidth. This pulsing behavior is what gives this architecture its name. [16].

This description sounds like the computational dynamics of a spreadsheet in Manual Mode. Manual mode is a calculation option that – in Excel – is specified with the Calculation Option dialog. Some of the major features of Manual Calculation in Excel that are exploited by Visual Backpropagation are reviewed in the Appendix. (Unfortunately, most spreadsheet systems do not support circular formula resolution in Manual Mode.) The basic concept is: calculation in Manual Mode proceeds in a systolic style from left-to-right and then top-down [17]. Microsoft calls this "Row Major Order" calculation [18]. We now put all this together.

### 3.1 Spreadsheet Regions for Visual Backpropagation

We need two regions to embed the backpropagation formulas: we call them Region A and Region B. Region A – always arranged above Region B – incorporates the initialization of random weights. We sometimes call Region A the predictor region and region B the corrector region (these terms are inspired from certain algorithms in numerical analysis). Both regions incorporate named arrays for the targets, inputs, outputs, and, and deltas.
For Region A define the following arrays. The array for the inputs:

$$\mathbf{inpA:}((n+1)\times 1) \quad .$$

Next, define the weights and outputs: starting from the input layer: for $h=1,2,\cdots,h,\cdots,q$ outputs:

$$\begin{array}{ll}
\mathbf{wA}_1:(m_1\times(n+1)); & \mathbf{outA}_1:((m_1+1)\times 1) \\
\mathbf{wA}_2:m_2\times(m_1+1); & \mathbf{outA}_2:((m_2+1)\times 1) \\
\ldots. & \ldots. \\
\mathbf{wA}_h:(m_h\times(m_{h-1}+1)); & \mathbf{outA}_h:((m_h+1)\times 1);\ \ h=1,..,q-1 \\
\ldots. & \ldots. \\
\mathbf{wA}_q:(m_q\times(m_{q-1}+1)); & \mathbf{outA}=\mathbf{outA}_q:(m_q\times 1)
\end{array}$$

Represent the deltas starting from the output layer: for $h=q,\cdots h,\cdots,2,1$

$$\begin{array}{l}
\boldsymbol{\delta}\mathbf{A}:(m_q\times 1):=\boldsymbol{\delta}\mathbf{A}_q:(m_q\times 1) \\
\boldsymbol{\delta}\mathbf{A}_h:(m_h\times 1) \\
\boldsymbol{\delta}\mathbf{A}_2:(m_2\times 1) \\
\boldsymbol{\delta}\mathbf{A}_1:(m_2\times 1)
\end{array}$$

Define similar arrays for Region B:

$$\mathbf{inpB:}((n+1)\times 1);$$

$$\begin{array}{ll}
\mathbf{wB}_1:(m_1\times(n+1)); & \mathbf{outB}_1:((m_1+1)\times 1) \\
\mathbf{wB}_2:m_2\times(m_1+1); & \mathbf{outB}_2:((m_2+1)\times 1) \\
\ldots & \ldots \\
\mathbf{wB}_h:(m_h\times(m_{h-1}+1)); & \mathbf{outB}_h:((m_h+1)\times 1);\ \ h=1,..,q-1 \\
\mathbf{wB}_q:(m_q\times(m_{q-1}+1)); & \mathbf{outB}=\mathbf{outB}_q:(m_q\times 1) \\
\boldsymbol{\delta}\mathbf{B}:(m_q\times 1):=\boldsymbol{\delta}\mathbf{B}_q:(m_q\times 1) & \\
\boldsymbol{\delta}\mathbf{B}_h:(m_h\times 1),\ h=q-1,\cdots,1 & \\
\boldsymbol{\delta}\mathbf{B}_2:(m_2\times 1) & \\
\boldsymbol{\delta}\mathbf{B}_1:(m_2\times 1) &
\end{array}$$

To provide for initialization: define a parameter $ru$ that is set to 0 for initialization and 1 for gradient search. We assume there is a function called $random()$ that assigns initial values. For example, in Excel, we can choose $random()$ to be `=rand()` – this returns a random decimal number between 0 and 1; more commonly we choose $random()$ to be `2*rand() -1` – this returns a random decimal number between -1 and 1. The choice is a design option.

For Region A, define and specify (order is significant) the following formulas:

1A. Define and specify a sample record of inputs and targets as a column vector dependent on spreadsheet iteration. The inputs and corresponding targets are selected by formulas (involving the OFFSET worksheet function) from a named spreadsheet region (This is discussed in the Appendix).

2A. Define and specify the weights and outputs as array formulas in row major order from left to right. For the first (input) layer $h = 1$:

$$\begin{cases} \mathbf{wA}_1 : (m_1 \times (n+1)) := if\left(ru = 0, random(), \mathbf{wB}_1 + \eta_1 \cdot (\mathbf{inpB})^T \otimes \boldsymbol{\delta}\mathbf{B}_1\right) \\ \mathbf{outA}_1 : ((m_1+1)\times 1) := \begin{pmatrix} f_1(\mathbf{wA}_1 \bullet \mathbf{inpA}) : (m_1 \times 1) \\ 1 \end{pmatrix} \end{cases}$$

In general, for $h = 1,..,q-1$ (where $n = m_0$ ):

$$\begin{cases} \mathbf{wA}_h : (m_h \times (m_{h-1}+1)) := if\left(ru = 0, random(), \mathbf{wB}_h + \eta_h \cdot (\mathbf{outB}_{h-1})^T \otimes \boldsymbol{\delta}\mathbf{B}_h\right) \\ \mathbf{outA}_h : ((m_h+1)\times 1) := \begin{pmatrix} f_h(\mathbf{wA}_h \bullet \mathbf{outA}_{h-1}) : (m_h \times 1) \\ 1 \end{pmatrix} \end{cases}$$

For last (output) layer $h = q$

$$\begin{cases} \mathbf{wA}_q : (m_q \times (m_{q-1}+1)) := if\left(ru = 0, random(), \mathbf{wB}_q + \eta_q \cdot (\mathbf{outB}_{q-1})^T \otimes \boldsymbol{\delta}\mathbf{B}_q\right) \\ \mathbf{outA} = \mathbf{outA}^q : (m_q \times 1) := f_q(\mathbf{wA}_q \bullet \mathbf{outA}_{q-1}) : (m_1 \times 1) \end{cases}$$

Note that the weight updates for Region A depend on results in Region B.

3A. Define and specify the deltas as array formulas in row major order from lefty to right:
For last layer $h = q$

$$\boldsymbol{\delta}\mathbf{A} : (m_q \times 1) := \boldsymbol{\delta}\mathbf{A}_q : (m_q \times 1) := (\mathbf{targA} - \mathbf{outA}) \odot \left( f_q{}'(\mathbf{z})\big|_{\mathbf{z}=\mathbf{wA}_q \bullet \mathbf{outA}_{q-1}} \right) : (m_q \times 1)$$

For $h = q-1,\cdots,1$

$$\boldsymbol{\delta}\mathbf{A}_h : (m_h \times 1) := \left( (\mathbf{wA}_{h+1})^T \bullet \boldsymbol{\delta}\mathbf{A}_{h+1} : (m_h \times 1) \right) \odot \left( f_h{}'(\mathbf{z})\big|_{\mathbf{z}=\mathbf{wA}_h \bullet \mathbf{outA}_{h-1}} \right); h = q-1,\cdots,1$$

Note that the updates for the outputs and deltas in Region A depend on results in Region A.

We similarly specify computation for Region B as follows:

1B. Define and specify a sample record of inputs and targets as a column vector dependent on spreadsheet iteration. The inputs and corresponding targets are selected by formulas (involving the OFFSET worksheet function) from a named spreadsheet region (This is discussed in the Appendix).

2B. Define and specify the weights and outputs as array formulas in row major order from left to right. For the first (input) layer $h=1$:

$$\begin{cases} \mathbf{wB}_1 : (m_1 \times (n+1)) := \mathbf{wA}_1 + \eta_1 \cdot \left(\mathbf{inpA}\right)^T \otimes \boldsymbol{\delta}\mathbf{A}_1 \\ \mathbf{outB}_1 : ((m_1+1)\times 1) := \begin{pmatrix} f_1(\mathbf{wB}_1 \bullet \mathbf{inpB}) : (m_1 \times 1) \\ 1 \end{pmatrix} \end{cases}$$

For $h=1,..,q-1$

$$\begin{cases} \mathbf{wB}_h : (m_h \times (m_{h-1}+1)) := \mathbf{wA}_h + \eta_h \cdot \left(\mathbf{outA}_{h-1}\right)^T \otimes \boldsymbol{\delta}\mathbf{A}_h \\ \mathbf{outB}_h : ((m_h+1)\times 1) := \begin{pmatrix} f_h(\mathbf{wB}_h \bullet \mathbf{outB}_{h-1}) : (m_h \times 1) \\ 1 \end{pmatrix} \end{cases}$$

For output

$$\begin{cases} \mathbf{wB}_q : (m_q \times (m_{q-1}+1)) := \mathbf{wA}_q + \eta_q \cdot \left(\mathbf{outA}_{q-1}\right)^T \otimes \boldsymbol{\delta}\mathbf{A}_q \\ \mathbf{outB} = \mathbf{outB}_q : (m_q \times 1) := f_q(\mathbf{wB}_q \bullet \mathbf{outB}_{q-1}) : (m_1 \times 1) \end{cases}$$

Note that the weight updates for Region B depend on results in Region A.

3B. Define and specify the deltas as array formulas in row major order from left to right:

$$\boldsymbol{\delta}\mathbf{B} : (m_q \times 1) := \boldsymbol{\delta}\mathbf{B}_q : (m_q \times 1) := \left(\mathbf{targB} - \mathbf{outB}\right) \odot \left( f_q{}'(\mathbf{z})\Big|_{\mathbf{z}=\mathbf{wB}_q \bullet \mathbf{outB}_{q-1}} \right) : (m_q \times 1)$$

For $h=q-1,\cdots,1$

$$\boldsymbol{\delta}\mathbf{B}_h : (m_h \times 1) := \left( \left(\mathbf{wB}_{h+1}\right)^T \bullet \boldsymbol{\delta}\mathbf{B}_{h+1} : (m_h \times 1) \right) \odot \left( f_h{}'(\mathbf{z})\Big|_{\mathbf{z}=\mathbf{wB}_h \bullet \mathbf{outB}_{h-1}} \right); \; h=q-1,\cdots,1$$

Note that the updates for the outputs and deltas in Region B depend on results in Region B.

Region A incorporates initialization when the spreadsheet name $ru := 0$; Region B always incorporates the values in Region A for weight updating. By left-right top-down semantics, these Region A values are available. Conversely, when $ru := 1$, Region A always incorporates the already computed values in Region B for weight updating: by left-right top-down semantics, these Region B values are available (see the Appendix for examples).

In practice, first initialize the set of weights in Region A by setting iteration to 1 (on Excel Options: Formulas: Calculation Options); setting $ru := 0$; and pressing Shift-F9 calculates the active worksheet (only!) and propagates the weights with random values. These random weights are matrix multiplied with the selected input record to compute and propagate values to the corresponding dependent outputs. After the last output is computed we compute the deltas. The initial weights are propagated to Region B. After initialization, "learning" occurs by setting iteration to a larger number – typically a multiple of the total number of record samples. After setting $ru := 1$; and pressing Shift-F9, the spreadsheet calculates all formulas in the active worksheet and propagates the weights, outputs, and delta values associated with backpropagation gradient search. After one iteration Excel uses these backpropagation weights and inputs to compute and propagate gradient search values to the corresponding dependent outputs and deltas across Region A, then to Region B, and back to Region A. The backpropagation stops when the iteration limit is reached.

The following shows a simple example of the regression problem of Figure 2.5 (example spreadsheets are posted in [8]). As observed in Section 2, this problem is not linearly separable. We used Visual Backpropagation to implement a 2-2-2-2 network with tanh as the activation function for all layers. After about 1000 iterations (250 times through all samples, or 250 "epochs") the computed weights are

| w_1 | | |
|---|---|---|
| 1.014045 | 1.034723669 | -1.24132 |
| 0.962683 | 0.887954178 | -0.49682 |

| w_2 | | |
|---|---|---|
| 1.774492 | 0.712279 | -0.79761 |
| 1.374877 | 1.563494 | 0.837633 |

| w_3 | | |
|---|---|---|
| -1.40842 | 1.394039 | -0.3026 |
| 1.388176 | -0.24171 | 1.184837 |

The output values compared to the targets are:

| targ1 | targ2 |
|---|---|
| 0 | 0 |
| 1 | 0 |
| 1 | 0 |
| 0 | 1 |

| out1 | out2 |
|---|---|
| 0.004604 | -0.00048138 |
| 0.950528 | -0.00218611 |
| 0.952855 | -0.0010876 |
| 0.006799 | 0.964574872 |

Clearly the nonlinear network results in a much better approximation than the naïve linear regression with the same inputs: the sum of the squared errors are less than 0.005 (targ1) and 0.0013 (targ2). Adding a third input to the linear regression – for example, the nonlinear combination x1*x2 – results in an even better (exact) approximation than the nonlinear network with two inputs. But is this addition biased? (The extra input is actually targ2).

Let's inspect the spreadsheet that computes this Visual Backpropagation [8]. First, note that the worksheet names in bold are defined by the DEFINE NAME (or Name Manager) utility. Circular formula semantics are reviewed in the Appendix.

Figure 3.1 shows the initial state after the definition of array region names and insertion of array formulas. The first formula in cell K5 is an iteration counter (a circular formula), used for information purposes. Cells K6:K7 define two numbers that help select two sample records. For 4 samples, these numbers range between 0 and 3. These are also circular formulas that use the MOD function. The training sample selected for an iteration weight update for Region A is in E11:H11; for Region B the sample is in E12:H12. Both ranges use the OFFSET function to select the values from the Training Data in E6:H9. Cell K11 defines the learning rate eta: here, set to 0.1 for all layers. Cell K14 defines the **ru** parameter: when set to zero, weights are randomized; otherwise, weights are updated according to the backpropagation update rule.

Region A and Region B are presented in the thick box surrounding B16:U27. This sheet shows the initialization so **ru** = 0 and iteration is set to 1. We proceed in left-to-right top-down semantics.

In Region A:
The targets and inputs **inpA** and **targA** are transposed from the values in E11:H11.
Set weights **w_1A** to random values.
Compute the first layer output **out_1A** with **inpA**, **w_1A** and a tanh function map.
Set weights **w_2A** to random values.
Compute the second layer output **out_2A** with **out_1A**, **w_2A** and a tanh function map.
Set weights **w_3A** to random values.
Compute the last (third) layer output **outA** with **out_2A**, **w_3A** and a tanh function map.
Compute delta **delA** for the last (third) output with **targA** and **outA**.
Compute delta **del_2A** for the second output with **delA, w_3A** and **out_2A**.
Compute delta **del_1A** for the first output with **del_2A, w_2A** and **out_1A**.

In Region B:
The targets and inputs **inpA** and **targA** are transposed from the values in E11:H11.
Update weights **w_1B** via backpropagation formulas with **w_1A**, **eta**, **inpA**, and **delA**.
Compute the first layer output **out_1B** with **inpB**, **w_1B** and a tanh function map.
Update weights **w_2B** via backpropagation formulas with **w_2A**, **eta**, **out_1A**, and **del_2A**.
Compute the second layer output **out_2A** with **out_1A**, **w_2A** and a tanh function map.
Update weights **w_3B** via backpropagation formulas with **w_3A**, **eta**, **out_2A**, and **delA**.
Compute the last (third) layer output **outB** with **out_2B**, **w_3B** and a tanh function map.
Compute delta **delB** for the last (third) output with **targB** and **outB**.
Compute delta **del_2B** for the second output with **delB, w_3B** and **out_2B**.
Compute delta **del_1B** for the first output with **del_2B, w_2B** and **out_1B**.

Figure 3.2 shows this sheet with **ru** = 1 and iteration set to 1000. The worksheet shows the state after learning with 1002 iterations (about 250 epochs). We proceed in left-to-right top-down semantics.

In Region A: (Changes from above indicatesd with *)
The targets and inputs **inpA** and **targA** are transposed from the values in E11:H11.
*Update weights **w_1A** via backpropagation formulas with **w_1B**, **eta**, **inpB**, and **delB**.
Compute the first layer output **out_1A** with **inpA**, **w_1A** and a tanh function map.
*Update weights **w_2A** via backpropagation formulas with **w_2B**, **eta**, **out_1B**, and **del_2B**.
Compute the second layer output **out_2A** with **out_1A**, **w_2A** and a tanh function map.
*Update weights **w_3A** via backpropagation formulas with **w_3B**, **eta**, **out_2B**, and **delB**.
Compute the last (third) layer output **outA** with **out_2A**, **w_3A** and a tanh function map.
Compute delta **delA** for the last (third) output with **targA** and **outA**.
Compute delta **del_2A** for the second output with **delA, w_3A** and **out_2A**.
Compute delta **del_1A** for the first output with **del_2A, w_2A** and **out_1A**.

In Region B:
Semantics of computation identical to above with **ru**=0.

**Figure 3.1. Visual Backpropagation initiation. Random weight selection (ru=0).**

| | C | D | E | F | G | H | I | J | K | L | M | N | O | P | Q | R | S | T | U |
|---|---|---|---|---|---|---|---|---|---|---|---|---|---|---|---|---|---|---|---|
| 3 | | | | | | | | | 1 epoch= | 4 iterations | | | | | | | | | |
| 4 | *Sample Training Data* | | | | | | | *Iteration Counter **itc** : selects training vectors **inp** and **targ** from data in **TrData** sequentially.* | | | | | | | | | | | |
| 5 | | **TrData** | *inp1* | *inp2* | *targ1* | *targ2* | | | 2 | | | | | | | | | | |
| 6 | | *0* | 0 | 0 | 0 | 0 | | **itc** | 1 | | | | | | | | | | |
| 7 | *4 samples* | *1* | 0 | 1 | 1 | 0 | | **itcp1** | 2 | | | | | | | | | | |
| 8 | | *2* | 1 | 0 | 1 | 0 | | *Samples selected cycle sequentially through database.* | | | | | | | | | | | |
| 9 | | *3* | 1 | 1 | 0 | 1 | | | | | | | | | | | | | |
| 10 | | | | | | | | Learning rate parameter | | | | | | | | | | | |
| 11 | *training sample A* | | 0 | 1 | 1 | 0 | | **eta** | 0.1 | | | | | | | | | | |
| 12 | *training sample B* | | 1 | 0 | 1 | 0 | | | | | | | | | | | | | |
| 13 | | | | | | | | *Initialization: set **ru** =0; set iterations :=1 (Calculation Options); Calculate Sheet (Shift-F9) and see the random weights* | | | | | | | | | | | |
| 14 | *Define regions (bold) for each state A and B* | | | | | | | **ru** | 0 | *Training with BackProp: set **ru** =1; set iterations :=1000; Calculate Sheet (Shift-F9) and see the weights cc* | | | | | | | | | |
| 15 | *per layer:* | 2 | | | | 2 | | | | 2 | | | | 2 | | | | | |
| 16 | | | | | | | | | | | | | | | | | | | |
| 17 | **targA** | **inpA** | **w_1A** | | (bias) | **out_1A** | **w_2A** | | (bias) | **out_2A** | **w_3A** | | (bias) | **outA** | | **delA** | **del_2A** | **del_1A** | |
| 18 | 1 | 0 | *0.790365* | *0.713147* | 0.501897 | 0.83818612 | *0.96* | *0.29* | *0.85* | 0.954383 | *0.150753* | *0.657625* | *0.460321* | 0.821464 | | 0.058059139 | -0.007441866 | -0.00336605 | |
| 19 | 0 | 1 | *0.422969* | *0.884237* | 0.154611 | 0.777432481 | *0.23* | *0.7* | *0.51* | 0.847149 | *0.804015* | *0.891733* | *0.188289* | 0.936777 | | -0.11470699 | -0.018099826 | -0.00585289 | |
| 20 | | 1 | | | | 1 | | | | 1 | | | | | | | | | |
| 21 | | | | | | | | | | | | | | | | | | | |
| 22 | | | | | | | | | | | | | | | | | | | |
| 23 | **targB** | **inpB** | **w_1B** | | (bias) | **out_1B** | **w_2B** | | (bias) | **out_2B** | **w_3B** | | (bias) | **outB** | | **delB** | **del_2B** | **del_1B** | |
| 24 | 1 | 1 | *0.790365* | *0.712811* | 0.50156 | 0.859629897 | *0.96* | *0.29* | *0.85* | 0.949193 | *0.156294* | *0.662544* | *0.466127* | 0.813374 | | 0.063158719 | -0.009466961 | -0.00407601 | |
| 25 | 0 | 0 | *0.422969* | *0.883651* | 0.154025 | 0.520477066 | *0.23* | *0.7* | *0.51* | 0.788533 | *0.793067* | *0.882015* | *0.176818* | 0.925359 | | -0.13298376 | -0.028535728 | -0.01649612 | |
| 26 | | 1 | | | | 1 | | | | 1 | | | | | | | | | |
| 27 | | | | | | | | | | | | | | | | | | | |

**Figure 3.2. Visual Backpropagation after 1000 iterations (ru=1).**

| | C | D | E | F | G | H | I | J | K | L | M | N | O | P | Q | R | S | T | U |
|---|---|---|---|---|---|---|---|---|---|---|---|---|---|---|---|---|---|---|---|
| 3 | | | | | | | | 1 epoch= 4 iterations | | | | | | | | | | | |
| 4 | *Sample Training Data* | | | | | | | *Iteration Counter* ***itc*** *: selects training vectors* ***inp*** *and* ***targ*** *from data in* ***TrData*** *sequentially.* | | | | | | | | | | | |
| 5 | | **TrData** | *inp1* | *inp2* | *targ1* | *targ2* | | | 1002 | | | | | | | | | | |
| 6 | | *0* | 0 | 0 | 0 | 0 | | **itc** | 1 | | | | | | | | | | |
| 7 | *4 samples* | *1* | 0 | 1 | 1 | 0 | | **itcp1** | 2 | | | | | | | | | | |
| 8 | | *2* | 1 | 0 | 1 | 0 | | *Samples selected cycle sequentially through database.* | | | | | | | | | | | |
| 9 | | *3* | 1 | 1 | 0 | 1 | | | | | | | | | | | | | |
| 10 | | | | | | | | Learning rate parameter | | | | | | | | | | | |
| 11 | *training sample A* | | 0 | 1 | 1 | 0 | | **eta** | 0.1 | | | | | | | | | | |
| 12 | *training sample B* | | 1 | 0 | 1 | 0 | | | | | | | | | | | | | |
| 13 | | | | | | | | *Initialization: set* ***ru*** *=0; set iterations :=1 (Calculation Options); Calculate Sheet (Shift-F9) and see the random weights* | | | | | | | | | | | |
| 14 | *Define regions (bold) for each state A and B* | | | | | | | **ru** | 1 | *Training with BackProp: set* ***ru*** *=1; set iterations :=1000; Calculate Sheet (Shift-F9) and see the weights co* | | | | | | | | | |
| 15 | *per layer:* | *2* | | | | *2* | | | | *2* | | | | *2* | | | | | |
| 16 | | | | | | | | | | | | | | | | | | | |
| 17 | **targA** | **inpA** | **w_1A** | | (bias) | **out_1A** | **w_2A** | | (bias) | **out_2A** | **w_3A** | | (bias) | **outA** | | **delA** | **del_2A** | **del_1A** | |
| 18 | 1 | 0 | *1.288202* | *1.282724* | -0.34708 | 0.733214578 | *1.52* | *0.69* | *0.4* | 0.843603 | *1.195474* | *-1.37576* | *-0.25116* | 0.922935 | | 0.011420438 | 0.003854705 | 0.004729026 | |
| 19 | 0 | 1 | *1.085548* | *1.075834* | -1.52043 | -0.41744271 | *-0.6* | *1.7* | *0.42* | -0.61866 | *-0.10083* | *1.368692* | *0.929009* | -0.00282 | | 0.002816105 | -0.007319054 | -0.00807652 | |
| 20 | | 1 | | | | 1 | | | | 1 | | | | | | | | | |
| 21 | | | | | | | | | | | | | | | | | | | |
| 22 | | | | | | | | | | | | | | | | | | | |
| 23 | **targB** | **inpB** | **w_1B** | | (bias) | **out_1B** | **w_2B** | | (bias) | **out_2B** | **w_3B** | | (bias) | **outB** | | **delB** | **del_2B** | **del_1B** | |
| 24 | 1 | 1 | *1.288202* | *1.283197* | -0.34661 | 0.735954555 | *1.52* | *0.69* | *0.4* | 0.84643 | *1.196437* | *-1.37647* | *-0.25002* | 0.922568 | | 0.011527128 | 0.004072782 | 0.006880413 | |
| 25 | 0 | 0 | *1.085548* | *1.075027* | -1.52123 | -0.41006157 | *-0.6* | *1.7* | *0.42* | -0.61268 | *-0.1006* | *1.368518* | *0.929291* | 0.005684 | | -0.00568366 | -0.014769272 | -0.0185363 | |
| 26 | | 1 | | | | 1 | | | | 1 | | | | | | | | | |
| 27 | | | | | | | | | | | | | | | | | | | |

Figure 3.3 shows the Data Selection Formulas. The OFFSET formula array returns the record from TrData that is displaced itc rows down from the top for Region A,and displaced itcp1 rows down from the top for Region B. Data is selected at each iteration.

Figure 3.4 shows the formulas in the target, input, first layer weights and first outputs for Regions A and Region B. Note that these are basically a cut-and-paste of the formulas described above in Section 3.1.

Figure 3.5 shows the formulas in the second layer weights and second outputs in both Regions.

| | D | E | F | G | H | I | J | K |
|---|---|---|---|---|---|---|---|---|
| 4 | | | | | | | *Iteration* | |
| 5 | **TrData** | *inp1* | *inp2* | *targ1* | *targ2* | | | =K5+1 |
| 6 | *0* | 0 | 0 | 0 | 0 | | **itc** | =MOD(itc+1,4) |
| 7 | *1* | 0 | 1 | 1 | 0 | | **itcp1** | =MOD(itc+1,4) |
| 8 | *2* | 1 | 0 | 1 | 0 | | *Samples* | |
| 9 | *3* | 1 | 1 | 0 | 1 | | | |
| 10 | | | | | | | Learning | |
| 11 | *sample A* | =OFFSET(TrData,itc,) | =OFFSET(Tr[ | =OFFSET(Tr[ | =OFFSET(Tr[ | | **eta** | 0.1 |
| 12 | *sample B* | =OFFSET(TrData,itcp1,) | =OFFSET(Tr[ | =OFFSET(Tr[ | =OFFSET(Tr[ | | | |
| 13 | | | | | | | *Initializa* | |
| 14 | | | | | | | **ru** | 1 |

**Figure 3.3. Visual Backpropagation data selection formulas.**

| | C | D | E | F | G | H | |
|---|---|---|---|---|---|---|---|
| 17 | **targA** | **inpA** | **w_1A** | | bias | **out_1A** | |
| 18 | =TRANSPOSE(G11:H11) | =TRANSPOSE(E11:F11) | *=IF(ru=0,RAND(),w_1B+eta*(TRANSPOSE(inpB)*del_1B))* | *=IF(* | *=IF(* | =TANH(MMULT(w_1A,inpA)) | = |
| 19 | =TRANSPOSE(G11:H11) | =TRANSPOSE(E11:F11) | *=IF(ru=0,RAND(),w_1B+eta*(TRANSPOSE(inpB)*del_1B))* | *=IF(* | *=IF(* | =TANH(MMULT(w_1A,inpA)) | = |
| 20 | | 1 | | | | 1 | |
| 21 | | | | | | | |
| 22 | | | | | | | |
| 23 | **targB** | **inpB** | **w_1B** | | bias | **out_1B** | |
| 24 | =TRANSPOSE(G12:H12) | =TRANSPOSE(E12:F12) | *=w_1A+eta*(TRANSPOSE(inpA)*del_1A)* | *=w_* | *=w_* | =TANH(MMULT(w_1B,inpB)) | = |
| 25 | =TRANSPOSE(G12:H12) | =TRANSPOSE(E12:F12) | *=w_1A+eta*(TRANSPOSE(inpA)*del_1A)* | *=w_* | *=w_* | =TANH(MMULT(w_1B,inpB)) | = |
| 26 | | 1 | | | | 1 | |
| 27 | | | | | | | |

**Figure 3.4. Formulas for target, input, layer 1 weights and first outputs (Regions A and Region B).**

| | H | I | J | K | L | |
|---|---|---|---|---|---|---|
| 17 | **out_1A** | **w_2A** | | bias | **out_2A** | |
| 18 | =TANH(MM | *=IF(ru=0,RAND(),w_2B+eta*(TRANSPOSE(out_1B)*del_2B))* | *=IF(* | *=IF(* | =TANH(MMULT(w_2A,out_1A)) | = |
| 19 | =TANH(MM | *=IF(ru=0,RAND(),w_2B+eta*(TRANSPOSE(out_1B)*del_2B))* | *=IF(* | *=IF(* | =TANH(MMULT(w_2A,out_1A)) | = |
| 20 | 1 | | | | 1 | |
| 21 | | | | | | |
| 22 | | | | | | |
| 23 | **out_1B** | **w_2B** | | bias | **out_2B** | |
| 24 | =TANH(MM | *=w_2A+eta*(TRANSPOSE(out_1A)*del_2A)* | *=w_* | *=w_* | =TANH(MMULT(w_2B,out_1B)) | = |
| 25 | =TANH(MM | *=w_2A+eta*(TRANSPOSE(out_1A)*del_2A)* | *=w_* | *=w_* | =TANH(MMULT(w_2B,out_1B)) | = |
| 26 | 1 | | | | 1 | |

**Figure 3.5. Formulas for layer 2 weights and second outputs (Regions A and Region B).**

Figure 3.6 shows the formulas in the third layer weights and third (final) outputs in both Regions.

Figure 3.7 shows the formulas for the three deltas for Regions A and Region B. The formulas in R31:S32 show the computation of the 4-period (corresponding to the number of records in the sample data set) exponential moving average of the absolute errors. We can use this to determine if the errors are converging to zero as calculation proceeds.

| | L | M | N | O | P |
|---|---|---|---|---|---|
| 17 | **out_2A** | **w_3A** | | (bias) | **outA** |
| 18 | =TANH(M | *=IF(ru=0,RAND(),w_3B+eta*(TRANSPOSE(out_2B)*delB))* | *=IF(ru* | *=IF(ru* | =TANH(MMULT(w_3A,out_2A)) |
| 19 | =TANH(M | *=IF(ru=0,RAND(),w_3B+eta*(TRANSPOSE(out_2B)*delB))* | *=IF(ru* | *=IF(ru* | =TANH(MMULT(w_3A,out_2A)) |
| 20 | 1 | | | | |
| 21 | | | | | |
| 22 | | | | | |
| 23 | **out_2B** | **w_3B** | | (bias) | **outB** |
| 24 | =TANH(M | *=w_3A+eta*(TRANSPOSE(out_2A)*delA)* | *=w_3A* | *=w_3A* | =TANH(MMULT(w_3B,out_2B)) |
| 25 | =TANH(M | *=w_3A+eta*(TRANSPOSE(out_2A)*delA)* | *=w_3A* | *=w_3A* | =TANH(MMULT(w_3B,out_2B)) |
| 26 | 1 | | | | |

**Figure 3.6. Formulas for layer 3 weights and third (final) outputs (Regions A and Region B).**

| | R | S | T |
|---|---|---|---|
| 17 | **delA** | **del_2A** | **del_1A** |
| 18 | =(targA-outA)*(1-outA^2) | =MMULT(TRANSPOSE(w_3A),delA)*(1-out_2A^2) | =MMULT(TRANSPOSE(w_2A),del_2A)*(1-out_1A^2) |
| 19 | =(targA-outA)*(1-outA^2) | =MMULT(TRANSPOSE(w_3A),delA)*(1-out_2A^2) | =MMULT(TRANSPOSE(w_2A),del_2A)*(1-out_1A^2) |
| 20 | | | |
| 21 | | | |
| 22 | | | |
| 23 | **delB** | **del_2B** | **del_1B** |
| 24 | =(targB-outB)*(1-outB^2) | =MMULT(TRANSPOSE(w_3B),delB)*(1-out_2B^2) | =MMULT(TRANSPOSE(w_2B),del_2B)*(1-out_1B^2) |
| 25 | =(targB-outB)*(1-outB^2) | =MMULT(TRANSPOSE(w_3B),delB)*(1-out_2B^2) | =MMULT(TRANSPOSE(w_2B),del_2B)*(1-out_1B^2) |
| 26 | | | |
| 27 | | | |
| 28 | | | |
| 29 | | *4-Term Exponential Moving Averages (EMA):* | |
| 30 | targ-out | *EMA* | |
| 31 | =ABS(targB-outB) | =IF(S31:S32=0,R31:R32,0.4*R31:R32+0.6*S31:S32) | |
| 32 | =ABS(targB-outB) | =IF(S31:S32=0,R31:R32,0.4*R31:R32+0.6*S31:S32) | |

**Figure 3.7. Formulas: Output delta, Second Layer Output Delta, and First Level Output Delta. Exponential Moving average of output errors (Regions A and Region B).**

# 4 Visual Backpropagation on a Practical Problem

We show Visual Backpropagation solving the "Auto MPG" problem found in the UCI Machine Learning repository [6]. This example includes a data set a table of 398 sample records listing inputs (such as Horsepower, Weight, Model Year) and a single target: the miles per gallon (MPG). This example is used in the Tensorflow regression tutorial [7]; there, the topology is 9-64-64-1. The Tensorflow model has 9 inputs, 64 outputs in the first layer, 64 outputs in the second layer, and 1 output for the approximated MPG (miles per gallon). RELU activation functions are used for the middle two layers and a softmax activation [14] is used for output. Learning rate of 0.001 is used for all layers.

In the Visual Backpropagation model we use a slightly different model in order to investigate two questions:

(1) Do the large number of outputs make a difference?

To investigate this we configure a Visual Backpropagation spreadsheet with 9-50-30-1 topology; there are 40% fewer weights to learn (2061 vs 4865). We use the same inputs and the same learning rates.

(2) Does softmax make a difference?

Following Øland et al's critique of softmax activation [19], we use the identity function as the activation function on the last output layer and still keep the RELU activation functions on the first two layers.

We follow the same procedure as in the Tensorflow tutorial using heuristics recommended by Lecun et al [2]:

1. Validate data: make sure that all input and target values are numbers and otherwise "make sense" for back propagation. This task is easy for spreadsheets. There are 392 purely numerical records.
2. Divide data into an in-Sample region (for training) and an out-Sample region (for evaluation or tuning). The Tensorflow example uses 80% of the data for training (314 records); we use a larger set of 360 samples (90%). The remainder (78 records and 32 records respectively) are used for out-sample analysis. Data selection tasks are easy for spreadsheets.
3. Scale data using values from the in-Sample Region. Scaled data improves backpropagation convergence. We use the z-score scaling (subtract the mean from each value and divide by the standard deviation) used in the Tensorflow example. This task is easy for spreadsheets.
4. Train with backpropagation.
5. Investigate errors associated with the in-Sample region. This task is easy for spreadsheets.
6. Cross-Validation: Investigate errors associated with the out-Sample region. For tuning, this could involve further application of backpropagation.

We discuss each step in turn. Our Visual Backpropagation spreadsheets are available at [8].

## 4.1 Validate Data and Select in- and out-Sample Subsets

We read the data into car-mpg-scaling.xlsx (see Tab: orig-data). Column C contains the target mpg; columns D through L contain the data. In column N we add a computed field called IsNumber: the values are the array formulas

```
= IF(ISNUMBER(SUM(ABS(C7:L7))),"Yes","No") // M7
```

We copy this array formula down through all rows. The simple formula uses the ISNUMBER worksheet function to determine if the sum of the absolute values in the preceding columns is a number: these records are valid; if the formula returns “NO” then there are non-numeric values somewhere in the record. We use this field with the Advanced Filter utility to extract all valid records from the data set as shown in Figure 4.1.

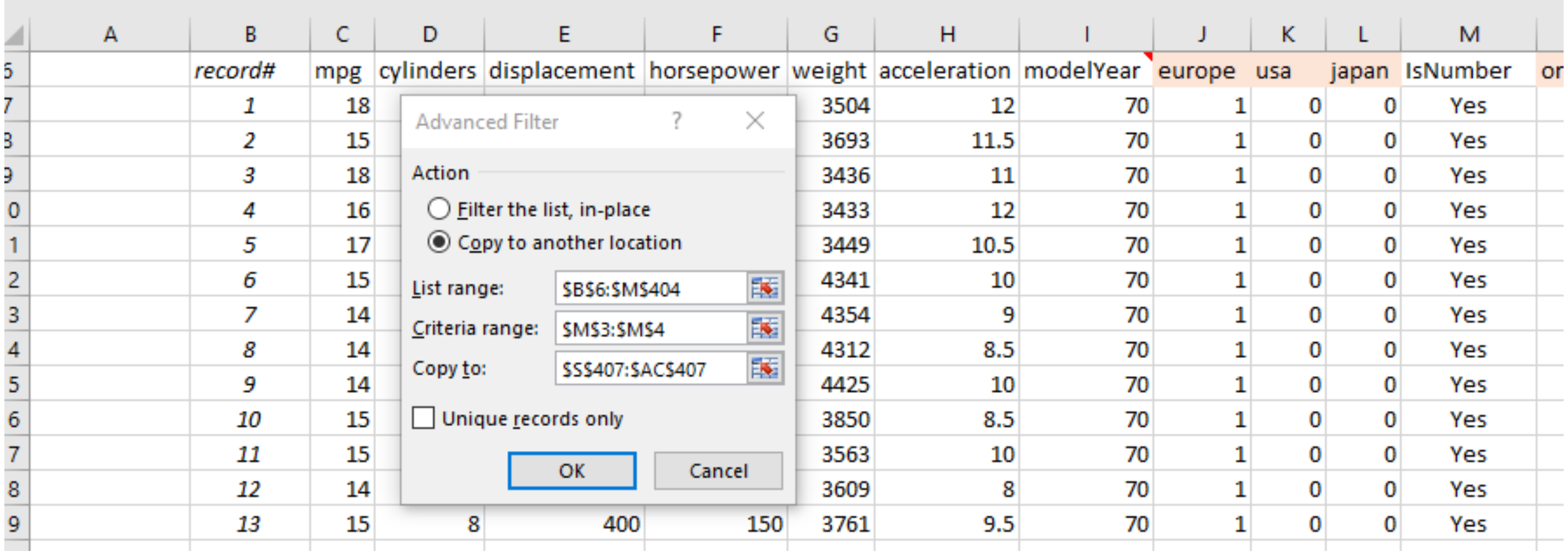

**Figure 4.1. Formulas: Output delta, Second Layer Output Delta, and First Level Output Delta. Exponential Moving average of output errors (Regions A and Region B).**

## 4.2 Validate Data and Select in- and out-Sample Subsets

From 398 records, 392 have only numbers (by inspection some have text like “?”). Select 360 records for in-Sample Training with backpropagation (rationale: 360 is a nice number). Note that more rigorous cross-validation scenarios can involve several out-Sample regions.

## 4.3 Scale data

Given a columns of field values $x$ create a rescaled value $y$. Train on these rescaled values: (all inputs and targets). Linear scales are simplest and invertible: $y = \alpha \cdot x + \beta$; and in the original units: $x = (y - \beta)/\alpha$. Note that to avoid bias, the scale parameters must be derived from the in-Sample data. In linear range scaling, suppose $A = \min(x)$ and $B = \max(x)$ in the original units over the in-Sample data. To scale between $a = \min(y)$ and $b = \max(y)$, then

$$\alpha = \frac{b-a}{B-A} \text{ and } \beta = \frac{B \cdot a - A \cdot B}{B-A}$$

Typically the rescaled $y$ values are between -1 and 1. Another linear scaling requires sample average $\mu$ and sample standard deviation $\sigma$. The “normalized z-score” scaling is

$$y = (x-\mu)/\sigma \text{ so } \alpha = 1/\sigma \text{ and } \beta = -\mu/\sigma .$$

Figure 4.2 shows the values computed for scaling for the Auto MPG problem.

| | C | D | E | F | G | H | I | J | K | L | M | N | O |
|---|---|---|---|---|---|---|---|---|---|---|---|---|---|
| 4 | | | mpg | cylinders | disp. | horsepow. | weight | **acceler.** | modelYear | europe | usa | japan | |
| 5 | | Max | 46.6 | 8 | 455 | 230 | 5140 | 24.8 | 81 | 1 | 1 | 1 | |
| 6 | | Min | 9 | 3 | 68 | 46 | 1613 | 8 | 70 | 0 | 0 | 0 | |
| 7 | | (Max+Min)/2 | 27.8 | 5.5 | 261.5 | 138 | 3376.5 | 16.4 | 75.5 | 0.5 | 0.5 | 0.5 | |
| 8 | | Mean | 22.75833 | 5.575 | 199.83 | 106.48611 | 3021.286 | 15.45333 | 75.45 | 0.622222 | 0.183333 | 0.194444 | |
| 9 | Standard Deviation | | 7.567275 | 1.727213 | 106.62 | 39.305708 | 864.2491 | 2.774745 | 3.36117539 | 0.484832 | 0.38694 | 0.395772 | |
| 10 | | (Max-Min)/4 | 9.4 | 1.25 | 96.75 | 46 | 881.75 | 4.2 | 2.75 | 0.25 | 0.25 | 0.25 | |
| 11 | | | | | | | | | | | | | |

**Figure 4.2. Values for scaling validated data.**

## 4.4 Training with Visual Backpropagation

For spreadsheet-based data, it is easy to see how linear regression performs. Using Excel's built-in functions LINEST and TREND we can determine weights, in-Sample errors, and out-Sample errors. (Note that LINEST outputs are ordered rightmost value first.) For Auto MPG, the average absolute in-Sample error in original units is 2.44 mpg; the average absolute out-Sample error in original units is 3.06 mpg. These values should be viewed as a baseline: we expect the nonlinear networks to outperform the standard linear values. The regression weights are shown in Figure 5.3.

| cylinders | disp. | horsepow | weight | acceler. | modelYea | europe | usa | japan | constant |
|---|---|---|---|---|---|---|---|---|---|
| -0.09957 | 0.285509 | -0.09792 | -0.72568389 | 0.018113 | 0.336016 | 0 | 0.125504 | 0.149563 | 2.29646E-15 |

**Figure 4.3. Linear Regression weights for Auto MPG.**

Note that the most significant linear factors (those with the largest magnitude of weights) are car weight and car year. This is intuitively appealing since weight adversely affects mileage; we also expect newer cars to be more efficient.

Setting up the 9-50-30-1 nonlinear network is straightforward: in our case it is a modification of the 2-2-2-2 network we used in the example shown in Figure 3.1 and Figure 3.2. First copy the scaled in-Sample data (360 records) to a visual backpropagation worksheet. Then define array regions for Region A and Region B. Then copy and paste the array formulas. To make the worksheet manageable, format the worksheet by hiding or displaying rows or columns. The final results after training are in vbp_car-mpg-RELU-9-50-30-1-optim.xlsx [8]. For example, Figure 4.4 shows the two training regions displaying weight rows 1,9,10,31,51 and columns 1 and 10 for the first layer weights; 1 and 51 for the second layer weights; and 1 and 31 for the third layer weights (see column G and row 374 for the row and column indices). Figure 4.5 shows the network topology with the Excel Trace Dependents utility. Figure 4.6 shows all weights formatted via "red-yellow-green" color scales in conditional formatting; this shows visually the highest and lowest weights.

On a 64-bit 2.7 GHz laptop, training speed is about 30 seconds per 100 epochs (360 samples). An exponential moving average of the errors was computed on the training sheet to give an indication of convergence: see cell DD485 on Figure 4.4. The Appendix (Section A.3) reviews some details on iterative (recursive) representations of exponential moving averages.

| | G | H | I | J | K | L | U | V | W | BU | BV | BW | DA | DB | DC | DD | DE | DF | DI |
|---|---|---|---|---|---|---|---|---|---|---|---|---|---|---|---|---|---|---|---|
| 373 | | | | | 9+1 x 1 | | | 50+1 x 1 | | | 30+1 | | | | | | | | |
| 374 | | | | | | 1 | 10 | | 1 | 51 | | 1 | 31 | | | | | | |
| 375 | | | | targA | inpA | w_1A | | out_1A | w_2A | | out_2A | w_3A | | outA | | delA | del_2A | del_1A | |
| 376 | 1 | cylinde | mpg | -0.63 | 1.404 | -0.57076 | 0.223 | 0 | 0.270562 | -1 | 2.49862 | 0.104 | 0.454 | -0.82 | | 0.1956 | 0.0203688 | 0 | 1 |
| 384 | 9 | japan | | | -0.491 | 0.25448 | -0.3 | 0.20776 | 0.215686 | 0.6 | 4.90352 | | | | | | -0.015344 | -0.06115 | 9 |
| 385 | 10 | | | | 1 | -0.00718 | -0.45 | 0 | -0.71185 | -1 | 0 | | | | | | 0 | 0 | 10 |
| 406 | 31 | | | | | -0.25956 | 0.137 | 0 | | | 1 | | | | | | | 0 | 31 |
| 426 | 51 | | | | | | | 1 | | | | | | | | | | | |
| 427 | | | | | | | | | | | | | | | | | | | |
| 428 | | | | | | | | | | | | | | | | | | | |
| 429 | | | | | | | | | | | | | | | | | | | |
| 430 | | | | | 9+1 x 1 | | | 50+1 x 1 | | | 30+1 | | | | | Linear output | | | |
| 431 | | | | | | 1 | 10 | | 1 | 51 | | 1 | 31 | | | | | | |
| 432 | | | | targB | inpB | w_1B | | out_1B | w_2B | | out_2B | w_3B | | outB | | delB | del_2B | del_1B | |
| 433 | 1 | cylinde | mpg | -0.89 | 1.404 | -0.57076 | 0.223 | 0 | 0.270562 | -1 | 1.94404 | 0.105 | 0.455 | -0.74 | | -0.149 | -0.015587 | 0 | 1 |
| 441 | 9 | japan | | | -0.491 | 0.2544 | -0.3 | 0.15258 | 0.215686 | 0.6 | 4.10333 | | | | | | 0.0115444 | 0.04666 | 9 |
| 442 | 10 | | | | 1 | -0.00718 | -0.45 | 0 | -0.71185 | -1 | 0 | | | | | | 0 | 0 | 10 |
| 463 | 31 | | | | | -0.25956 | 0.137 | 0 | | | 1 | | | | | | | 0 | 31 |
| 483 | 51 | | | | | | | 1 | | | | | | | 1-aL | 0.9945 | | | |
| 484 | | | | | | | | | | | | | | targ-Out | | EMA | | | |
| 485 | | | | | | | | | | | | | | 0.149 | | 0.18 | | | |

**Figure 4.4. Region A and Region B for Auto MPG [8] with hidden rows and columns.**

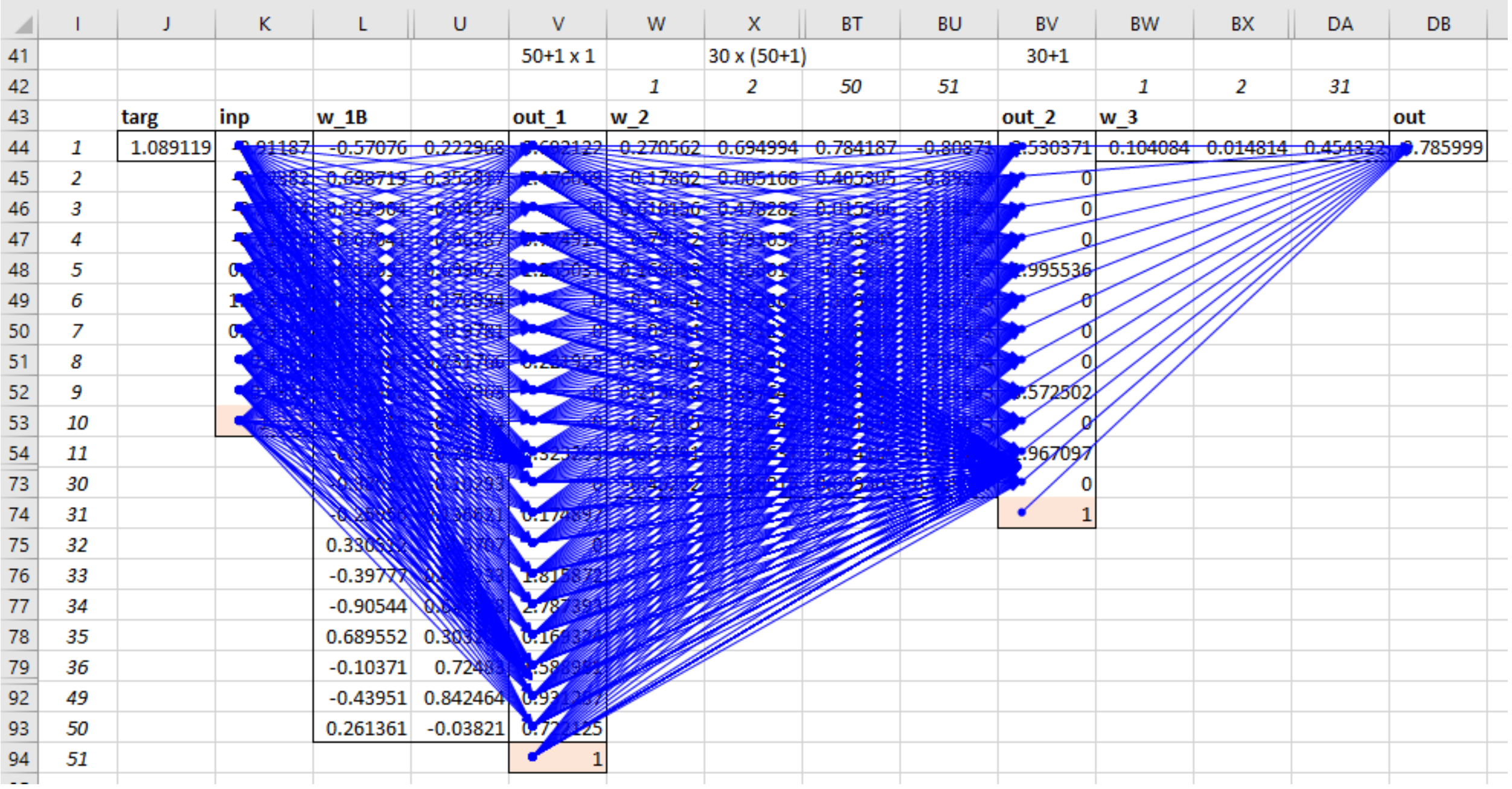

**Figure 4.5. Network topology for Auto MPG via the Excel Trace Dependents utility.**

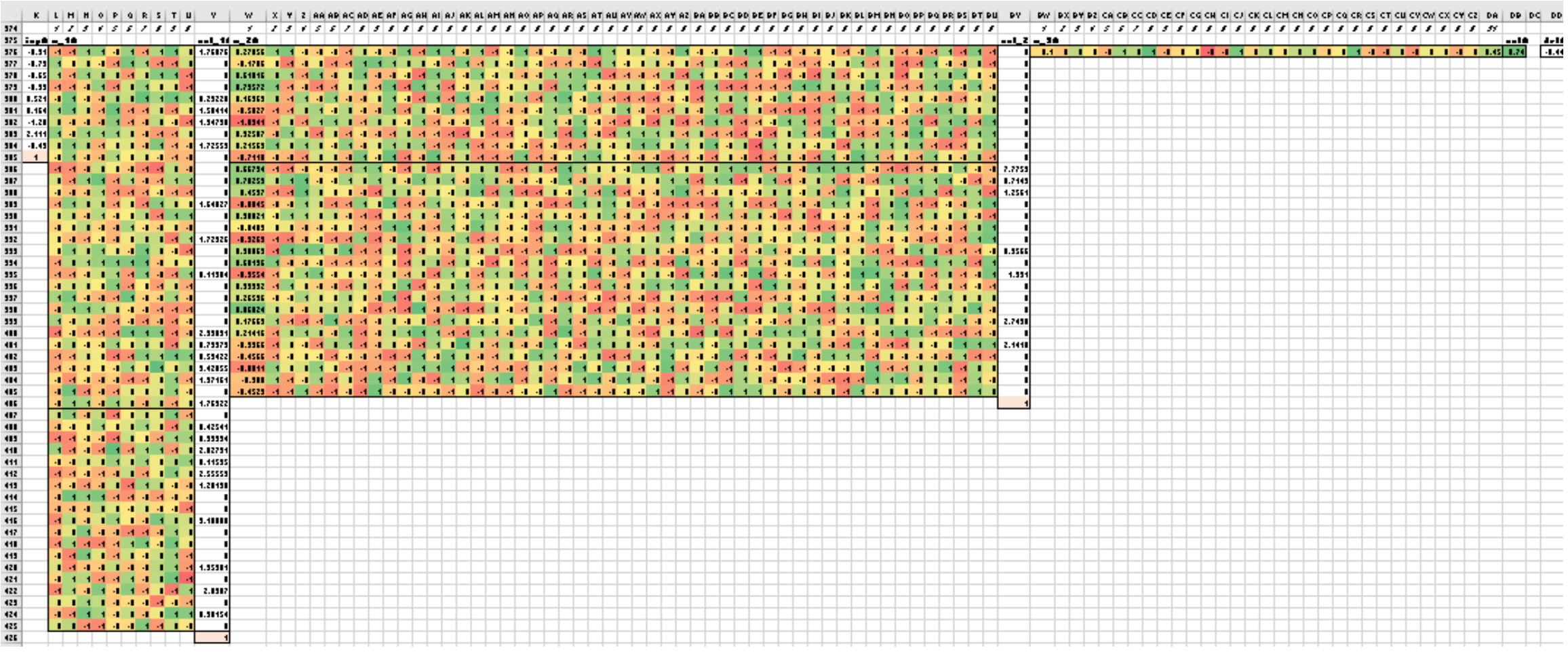

**Figure 4.6. The 3 weight regions for Auto MPG (formatted with red-yellow-green color scale) .**

## 4.5 Investigate errors associated with the in-Sample region

The charts in Figure 4.7 compare the results of our 9-50-30-1 network after about 80 epochs with the results from linear regression (Method III in Section 2.2). Both charts show that the 9-50-30-1 offers improvement over regression. Would more training or would different initial weights yield better results? Once the network is set up it is easy to try to answer this by restarting the iterations. We can easily change the learning parameters or activation functions as well in order to "tune" the 9-50-30-1 network.

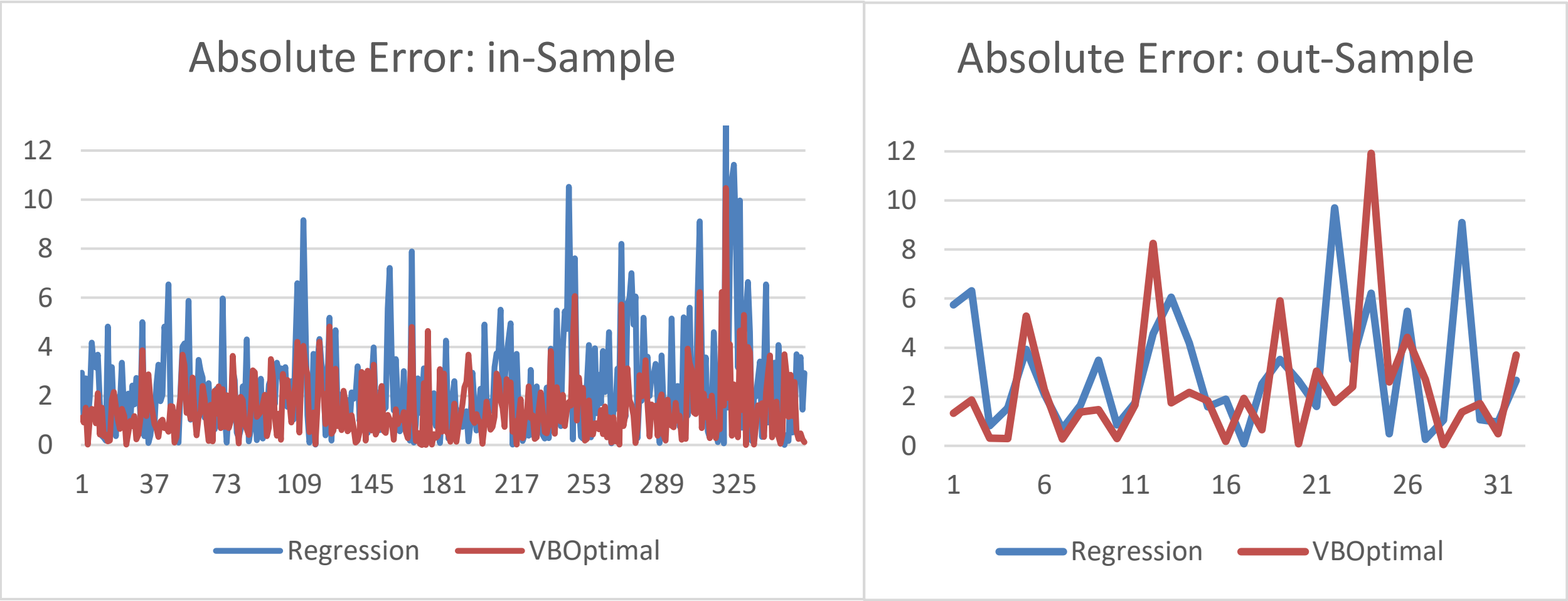

**Figure 4.7. Absolute Errors: 9-50-30-1 (RELU-RELU-Id; 80 epochs) vs. 9-1 (linear regression). The horizontal axis shows the sample number. Left: in Sample; Right: out-Sample.**

## 4.6 Cross-Validation: Investigate errors associated with the out-Sample region

Cross-validation has two meanings. In one interpretation, cross-validation provides estimates of the true model error. It answers the question: how well does a model predict? In some sense, looking at the one or more sets of out-sample data can provides an indication of predictability. The second interpretation of cross-validation provides a way to perform model tuning: the selection of model parameters that improve predictability. This interpretation seems to be prevalent in machine learning.

Here is an example of the second interpretation. Suppose at the end of each training epoch we use the weights to look at the out-sample errors. If the current out-sample error is smaller than the out-sample error of the previous training epoch, then save these weights; otherwise keep the previous weights. If we continue this procedure epoch-by-epoch, we should get a set of weights with the lowest out-sample error. Figure 4.8 shows a chart of the results and the "cross-validation procedure" written in VBA (whose only use is to control computation and save the results).

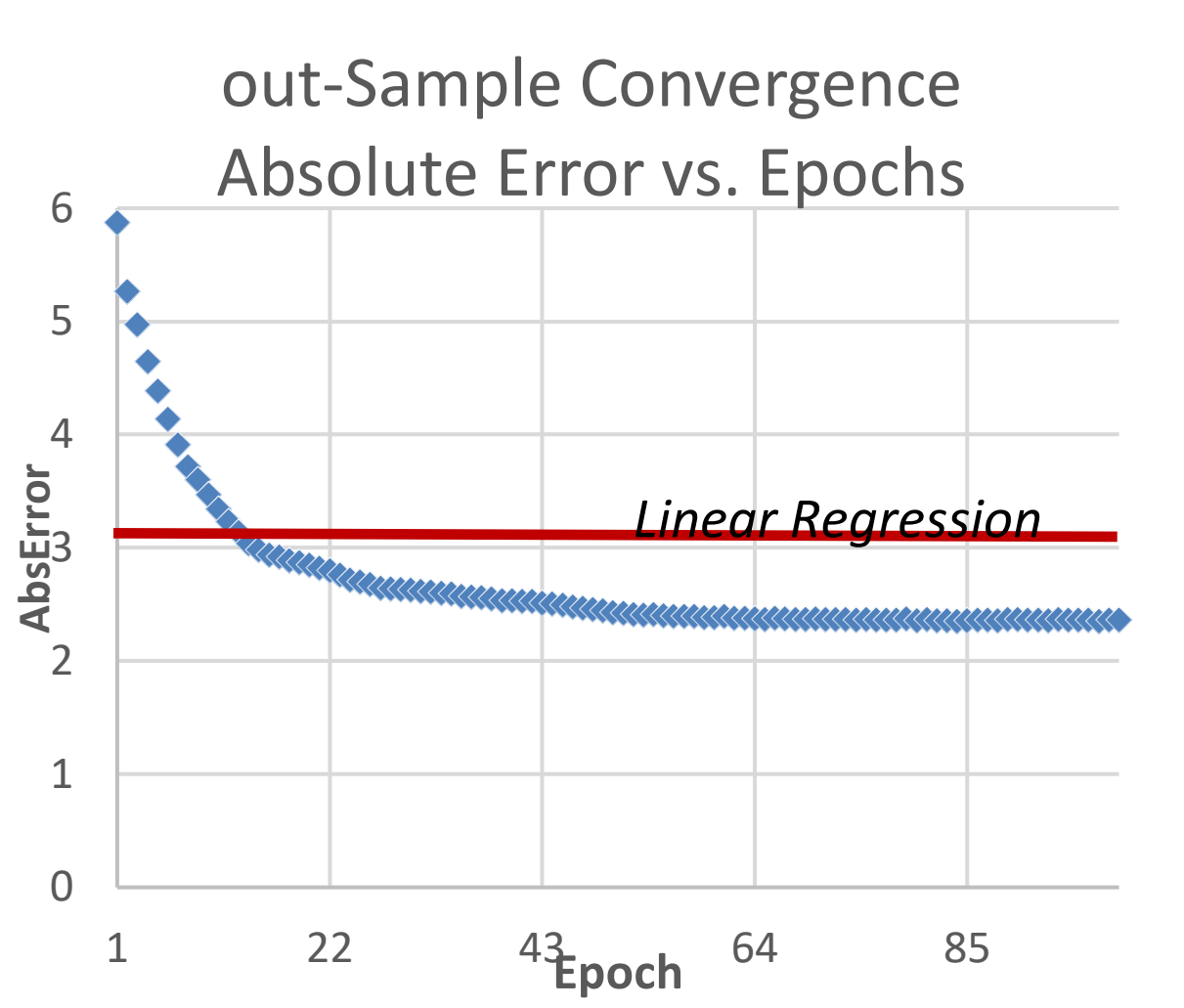

```
Sub lrn()
 For n = 1 To 100
  Worksheets("Training").Calculate
  Worksheets("out-Sample").Calculate
  Range("F135").Offset(n, 0).Value = n
  Range("G135").Offset(n,1).Value = Range("U132").Value
  If Range("U132").Value < Range("W132").Value Then
    Range("W132").Value = Range("U132").Value
    Range("X132").Value = n
   '' save workbook
    ActiveWorkbook.Save
  End If
 Next n
End Sub
```

**Figure 4.8. The use of cross-validation in model selection.**

As the number of epochs increase, the average absolute out-sample errors gradually decrease. We also show the improvement our 9-50-30-1 network over the average absolute out-sample errors over linear regression. Figure 4.9 summarizes the cross-validation findings for three models for the Auto MPG problem.

| Model | Avg. Absolute Error (mpg): in-Sample | Avg. Absolute Error (mpg): out-Sample |
|---|---|---|
| Linear Regression 9-1 | 2.44 | 3.06 |
| Visual Backpropagation 9-50-30-1 | 1.47 (epoch 84)<br>0.878 (epoch 1000) | 2.36 (epoch 84)<br>2.35 (epoch 118) |
| Tensorflow [7] 9-64-64-1 | 0.997 (epoch 995) | 1.95 (?) |

**Figure 4.9. Model Comparison: 9-1 vs. 9-50-30-1 vs. 9-64-64-1**

Note that the "best" average absolute in-sample and out-sample errors for the two nonlinear networks were obtained from weights that were tuned with the out-sample data. Many authors [20-22] have said that using single cross-validation procedure is for model tuning and for error assessment and estimation is problematic. At best the results are biased and at worst it can lead to under-estimation of the true error. The selection bias can be reduced if we set up another set of out-sample cross-validation data. (Further discussion of cross-validation is beyond the scope of this paper.)

## 5 Conclusions

Visual Backpropagation embeds a functional programming specification of the backpropagation algorithm within spreadsheets. This spreadsheet implementation is convenient and understandable for non-experts in machine learning and artificial intelligence. It is suitable for solving problems that are represented in spreadsheets.

One consequence of Visual Backpropagation spreadsheets is that they are pure formula-based computations. These spreadsheets can be readily shared. No software, programs, or libraries need to be downloaded and installed, called, or accessed. There are no security concerns or trust settings involving special macros: there are no macros.

A second consequence is that the backpropagation implementation is reasonably fast for interpretive languages: we know that built-in Excel functions are faster than the corresponding programs in VBA [12], and there are some indications that VBA is faster than Python (see e.g. [23-24]). Moreover, at least with Excel, we can easily enable multiprocessing speedup with (select "multi-threaded calculation" under "Advanced Options"). For this laptop, the option "Use all processors on this computer" was selected.

## Acknowledgements

I wish to acknowledge Dr. R. Down, S. Pennebaker, Dr. F. Hilf and W.P. Stahl for their encouragement and helpful conversations.

# APPENDIX: Computational Semantics of Manual Calculation

We review some of the major features of Manual Calculation that are exploited in Visual Backpropagation.

Figure A.1 shows the Excel Calculation Options Dialog. For Visual Backpropagation: Set the Workbook Calculation to Manual. Uncheck Recalculate before saving. Enable iterative calculation. When setting up regions: set Maximum Iterations to 1. When using Visual Backpropagation for training: set Maximum iterations to the number of records in the training data set. This means that when pressing <shift-F9> (Function key F9) – the Calculate Active Sheet command – the backpropagation algorithm iterates across the entire data set. This is called “1 epoch” in the machine learning literature.

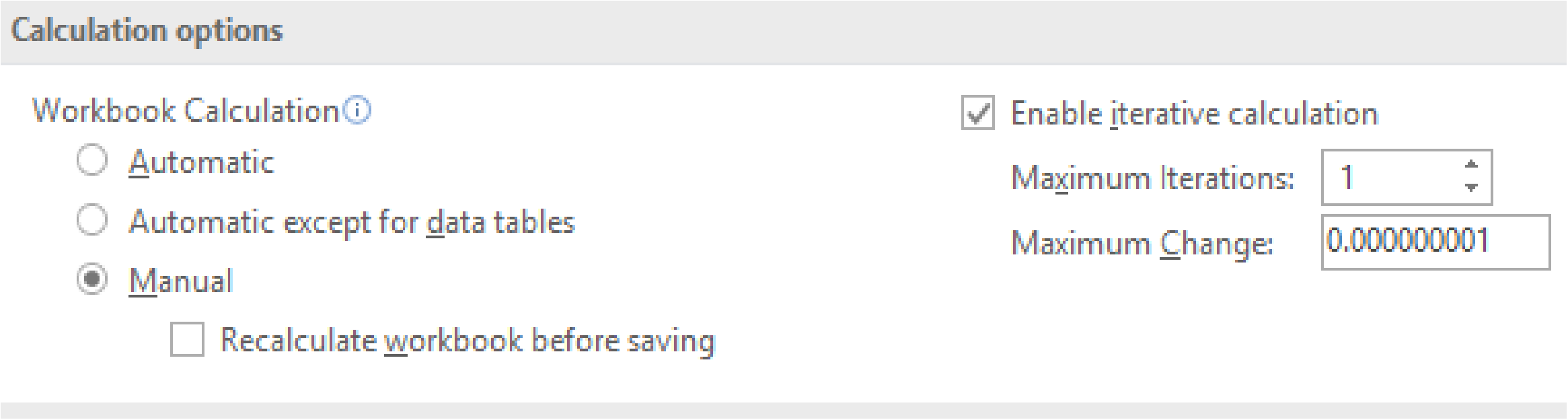

**Figure A.1. Excel Calculation Options Dialog.**

## A.1 Recursive Formulas

A circular formula is a formula that refers to itself. For Automatic Mode this normally indicates an error. In Manual mode circular formulas implement recursion. See Figure 3.2 for an example.

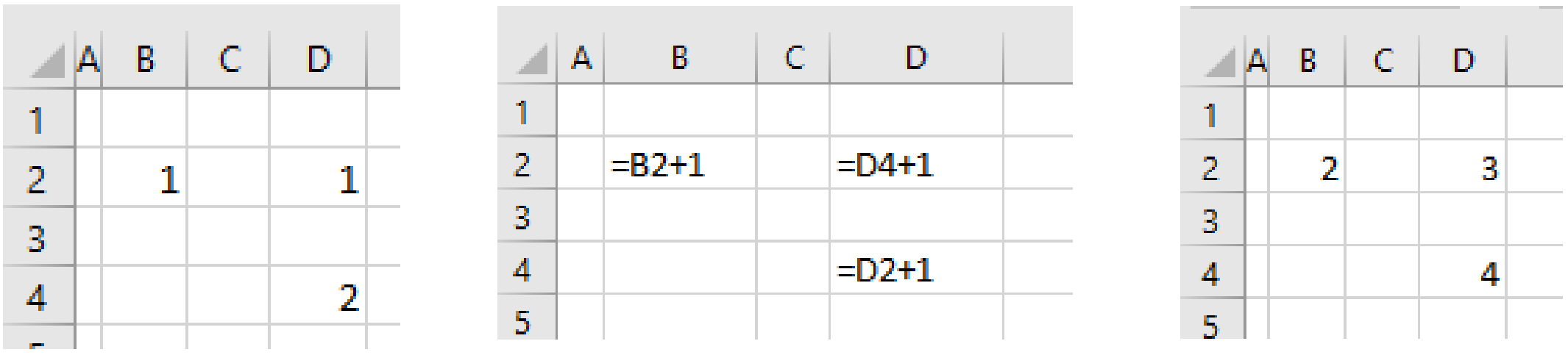

**Figure A.2. Circular formulas in cells B2, D2, and D4. Iteration is set to 1.**

In Figure A.2, the initial contents of these cells after the formulas are entered are shown in the left panel. The middle panel shows the formulas. Note that the default contents are zero. After formula entry, the contents of B2 is the previous value (zero) plus 1. The contents of D2 is the previous contents of D4 (zero) plus 1. The contents of D4 is the previous contents of D2 (1) plus 1. Now press <shift F9> to calculate this (the active) sheet. After 1 iteration, the right panel shows the values of these cells.

Calculation in Manual Mode proceeds in a systolic style from left-to-right and then top-down [17]. Microsoft calls this "Row Major Order" calculation [18]. Figure A.3 and A.4 show another example which demonstrates how previous states can be saved (this has been used in Visual Backpropagation to model autoregressive time series models (called recurrent networks in the machine learning literature).

| | C | D | E | F | G | H | I |
|---|---|---|---|---|---|---|---|
| 19 | *state:* | x(n-3) | x(n-2) | x(n-1) | **x(n)** | x(n) | x(n) |
| 20 | *value:* | 1 | 1 | 1 | **1** | 1 | 1 |
| 21 | | | | | | | |
| 24 | | *state:* | *value:* | | | | |
| 25 | | x(n-3) | 1 | | | | |
| 26 | | x(n-2) | 1 | | | | |
| 27 | | x(n-1) | 1 | | | | |
| 28 | | **x(n)** | 1 | | | | |
| 29 | | x(n) | 1 | | | | |
| 30 | | x(n) | 1 | | | | |

| | C | D | E | F | G | H | I |
|---|---|---|---|---|---|---|---|
| 19 | *state:* | x(n-3) | x(n-2) | x(n-1) | **x(n)** | x(n) | x(n) |
| 20 | *value:* | =E20 | =F20 | =G20 | **=G20+1** | =G20 | =H20 |
| 21 | | | | | | | |
| 24 | | *state:* | *value:* | | | | |
| 25 | | x(n-3) | =E26 | | | | |
| 26 | | x(n-2) | =E27 | | | | |
| 27 | | x(n-1) | =E28 | | | | |
| 28 | | **x(n)** | =E28+1 | | | | |
| 29 | | x(n) | =E28 | | | | |
| 30 | | x(n) | =E29 | | | | |

**Figure A.3. Left panel: initial state after formulas are entered.**
**Right panel: formulas. Note that G20 and E28 are the only circular formulas.**

| | C | D | E | F | G | H | I |
|---|---|---|---|---|---|---|---|
| 19 | *state:* | x(n-3) | x(n-2) | x(n-1) | **x(n)** | x(n) | x(n) |
| 20 | *value:* | 1 | 1 | 2 | **3** | 3 | 3 |
| 21 | | | | | | | |
| 24 | | *state:* | *value:* | | | | |
| 25 | | x(n-3) | 1 | | | | |
| 26 | | x(n-2) | 1 | | | | |
| 27 | | x(n-1) | 2 | | | | |
| 28 | | **x(n)** | 3 | | | | |
| 29 | | x(n) | 3 | | | | |
| 30 | | x(n) | 3 | | | | |

| | C | D | E | F | G | H | I |
|---|---|---|---|---|---|---|---|
| 19 | *state:* | x(n-3) | x(n-2) | x(n-1) | **x(n)** | x(n) | x(n) |
| 20 | *value:* | 1 | 2 | 3 | **4** | 4 | 4 |
| 21 | | | | | | | |
| 24 | | *state:* | *value:* | | | | |
| 25 | | x(n-3) | 1 | | | | |
| 26 | | x(n-2) | 2 | | | | |
| 27 | | x(n-1) | 3 | | | | |
| 28 | | **x(n)** | 4 | | | | |
| 29 | | x(n) | 4 | | | | |
| 30 | | x(n) | 4 | | | | |

**Figure A.4. Left: State after two presses to <shift-F9> to Calculate active sheet.**
**Right: State after another press: three presses total to SHIFT-F9 (Calculate Sheet).**

Note that the unshaded values in row 20 to the left of the circular formula are lagged. The colored values in column E that are above the circular formula are also lagged.

## A.2 Application: Selecting Data

We exploit these computational semantics to create formulas that dynamically select input records and corresponding target values from a dataset represented as a spreadsheet region. This helps enable "stochastic search" for visual backpropagation. Figure A.5 shows the spreadsheet regions and formulas. Define dataset TrData as the region E6:H9. There are 4 sample records. Cells K5, K6 and K7 contain circular formulas that always evaluate to be between 0 and 3: we define **itc** as K6 and **itcp1** as K7.

| | B | C | D | E | F | G | H | I | J | K |
|---|---|---|---|---|---|---|---|---|---|---|
| 3 | | | | | | | 100 epochs = | 400 iterations | | |
| 4 | | *Sample Training Data* | | | | | | | | |
| 5 | | | **TrData** | *inp1* | *inp2* | *targ1* | *targ2* | | | 1 |
| 6 | | | *0* | 0 | 0 | 0 | 0 | | **itc** | 1 |
| 7 | | *4 samples* | *1* | 0 | 1 | 1 | 0 | | **itcp1** | 2 |
| 8 | | | *2* | 1 | 0 | 1 | 0 | | | |
| 9 | | | *3* | 1 | 1 | 0 | 1 | | | |
| 10 | | | | | | | | | | |
| 11 | | *training sample A* | | 0 | 1 | 1 | 0 | | | |
| 12 | | *training sample B* | | 1 | 0 | 1 | 0 | | | |

| | D | E | F | G | H | I | J | K |
|---|---|---|---|---|---|---|---|---|
| 3 | | | | | 100 epochs = | 400 | | |
| 4 | | | | | | | | |
| 5 | **TrData** | *inp1* | *inp2* | *targ1* | *targ2* | | | =K5+1 |
| 6 | *0* | 0 | 0 | 0 | 0 | | **itc** | =MOD(itc+1,4) |
| 7 | *1* | 0 | 1 | 1 | 0 | | **itcp1** | =MOD(itc+1,4) |
| 8 | *2* | 1 | 0 | 1 | 0 | | | |
| 9 | *3* | 1 | 1 | 0 | 1 | | | |
| 10 | | | | | | | | |
| 11 | *training sample A* | =OFFSET(TrData,itc,) | =OFFSET(TrData,itc,) | =OFFSET(TrData,itc,) | =OFFSET(TrData,itc,) | | | |
| 12 | *training sample B* | =OFFSET(TrData,itcp1,) | =OFFSET(TrData,itcp1,) | =OFFSET(TrData,itcp1,) | =OFFSET(TrData,itcp1,) | | | |

**Figure. A.5. Dynamic Sample Selection: Top: values view. Bottom: formula view.**

Rows 11 and 12 contain array formulas with the OFFSET worksheet function. This function returns a reference to a region of a specified size, offset by rows and columns. At the end of the first iteration, note that itc=1 and itcp1=2. The OFFSET formulas use these values to select a single row of TrData displaced by itc rows and itcp1 rows. With these values the second and third rows of TrData are selected. With more iterations, subsequent rows are selected; the MOD function (which gives the remainder after division) ultimately resets itcp1 (and then itc) to 0.

Data can also shuffled before selected or can be randomly selected. Figure A.6 shows some examples of such formulas on a larger data set of 360 records.

| | |
|---|---|
| | =W365+1 |
| itc | =MOD(itc+181,360) |
| itcp1 | =MOD(itc+1,360) |
| | =W365+1 |
| itc | =MOD(itc+181,360) |
| itcp1 | =MOD(itc+31,360) |
| | =W365+1 |
| itc | =MOD(itc+RANDBETWEEN(0,359),360) |
| itcp1 | =MOD(itc+1,360) |

**Figure A.6. Shuffle and random selection of data.**

## A.3 Monitoring Absolute Error with Exponential Moving Averages

Given a sequence of values $y_1, y_2, ..., y_{n,...}$ the exponential moving average (EMA) is iteratively specified by

$$EMA_1 = y_1$$
$$EMA_{n+1} := \alpha \cdot y_{n+1} + (1-\alpha) \cdot EMA_n$$

We use this to approximate the simple moving average [19, 20] over the last $N$-periods. It can be shown that for

$$\alpha = 2/(N+1)$$

then the exponential moving average is a reasonable approximation to the $N$-period simple moving average, in that the last $N$ terms represent almost 90% of the total weight in the calculation. The representation is:

$$EMA_{n+1} := \left(\tfrac{2}{N+1}\right) \cdot y_{n+1} + \left(\tfrac{N-1}{N+1}\right) \cdot EMA_n$$

We use the exponential moving average to monitor the error for Visual Backpropagation training (e.g., see Figure 3.7). Figure A.7 shows the circular array formulas for a 4-period EMA.

| | J | K | L |
|---|---|---|---|
| 23 | | | *Absolute Error: 4-Term Exponential Moving Averages* |
| 24 | | targ-out | *EMA* |
| 25 | | =ABS(targ-out) | =IF(L25:L26=0,K25:K26,0.4*K25:K26+0.6*L25:L26) |
| 26 | | =ABS(targ-out) | =IF(L25:L26=0,K25:K26,0.4*K25:K26+0.6*L25:L26) |
| 27 | | | |

**Figure. A.7. A 4-period EMA for absolute errors.**

## A.4 Tabulating out-Sample Values

Suppose we want to tabulate the outputs of a complex set of formulas whose inputs are driven by a dynamic sample section as described above. For example, look at Figure A.8.

| | C | D | E | F | G | H |
|---|---|---|---|---|---|---|
| 5 | | Samples | *inp1* | *inp2* | *targ1* | *targ2* |
| 6 | | *0* | 0 | 0 | 0 | 0 |
| 7 | *! samples* | *1* | 0 | 1 | 1 | 0 |
| 8 | | *2* | 1 | 0 | 1 | 0 |
| 9 | | *3* | 1 | 1 | 0 | 1 |
| 10 | | | | | | |
| 11 | ***Sample#*** | =MOD(D11+1,4) | =OFFSET(Samples,D1 | =OFFSET(Samples,D11,) | =OFFSET(Samples,D: | =OFFSET(Samples,D: |
| 12 | | | | | | |

**Figure. A.8. Tabulating Data: Input Selection (current record number selected in D11).**

Here D11 contains the sample number for the input that cycles through the dataset. The two outputs we want tabulated are shown in Figure A.9.

| | B | C | D | E | F | G | H | I | J | K | L | M | N | O | P |
|---|---|---|---|---|---|---|---|---|---|---|---|---|---|---|---|
| 17 | | targ | inp | w_1 | | | out_1 | w_2 | | | out_2 | w_3 | | | out |
| 18 | | 1 | 0 | *1.014896* | *1.035574548* | -1.24065 | -0.20225 | *1.775053* | *0.713032* | *-0.79671* | -0.7116 | *-1.40833* | *1.393707* | *-0.30451* | 0.950126 |
| 19 | | 0 | 1 | *0.962825* | *0.888096135* | -0.49727 | 0.372071 | *1.375288* | *1.563712* | *0.837125* | 0.814678 | *1.388428* | *-0.24132* | *1.185454* | 0.000855 |
| 20 | | | 1 | | | | 1 | | | | 1 | | | | |
| 21 | | | | | | | | | | | | | | | |

**Figure. A.9. Tabulating Data: Output Selection (P18:P19).**

Formulas (for this example dataset of 4 records) that enable this are in Figure A.10. After 1 iteration, formulas in columns E and F ask if the sample number of the input (an integer between 0 and 3) equals to the cell label in the tabulated table (D30:D33) where the outputs are tabulated. The cells where the label equals the input sample number are changed to the corresponding output; otherwise the cell values in the tabulation are left unchanged. After 4 iterations the output values are all tabulated.

| | C | D | E | F | G | H | I | J | K | L |
|---|---|---|---|---|---|---|---|---|---|---|
| 28 | | ***Tabulate in-Sample Results*** | | | | | | | | |
| 29 | | *Sample#* | out1 | out2 | *targ1* | *targ2* | | | Absolute Errors | |
| 30 | | *0* | 0.00314 | -0.00052015 | 0 | 0 | | | 0.00314 | 0.00052 |
| 31 | | *1* | 0.952534 | -0.00115804 | 1 | 0 | | | 0.047466 | 0.001158 |
| 32 | | *2* | 0.942216 | 0.001913256 | 1 | 0 | | | 0.057784 | 0.001913 |
| 33 | | *3* | 0.004265 | 0.960234482 | 0 | 1 | | | 0.004265 | 0.039766 |
| 34 | | | | | | *Average:* | | | *0.028164* | *0.010839* |

| | C | D | E | F | G | H | I | J | K | L |
|---|---|---|---|---|---|---|---|---|---|---|
| 28 | | ***Tabulate*** | | | | | | | | |
| 29 | | *Sample#* | out1 | out2 | *targ1* | *targ2* | | | Absolute Errors | |
| 30 | | *0* | =IF($D$11=D30:D33,$P$18,E30:E33) | =IF($D$11=D30:D33,$P$19,F30:F33) | 0 | 0 | | | =ABS(G30:G33-E30:E33) | =ABS(H30:H33-F30:F33) |
| 31 | | *1* | =IF($D$11=D30:D33,$P$18,E30:E33) | =IF($D$11=D30:D33,$P$19,F30:F33) | 1 | 0 | | | =ABS(G30:G33-E30:E33) | =ABS(H30:H33-F30:F33) |
| 32 | | *2* | =IF($D$11=D30:D33,$P$18,E30:E33) | =IF($D$11=D30:D33,$P$19,F30:F33) | 1 | 0 | | | =ABS(G30:G33-E30:E33) | =ABS(H30:H33-F30:F33) |
| 33 | | *3* | =IF($D$11=D30:D33,$P$18,E30:E33) | =IF($D$11=D30:D33,$P$19,F30:F33) | 0 | 1 | | | =ABS(G30:G33-E30:E33) | =ABS(H30:H33-F30:F33) |
| 34 | | | | | | | | *Ave* | *=AVERAGE(K30:K33)* | *=AVERAGE(L30:L33)* |

**Figure. A.7. Tabulating values. Top: Values view. Bottom: Formulas view.**